\documentclass{zpx}

\usepackage{algorithm}
\usepackage{algorithmic}
\usepackage{amsmath}
\usepackage{amsfonts}
\usepackage{amsthm}
\usepackage{adjustbox}
\usepackage{xspace}

\newtheorem{proposition}{Proposition}
\newtheorem{corollary}{Corollary}

\makeatletter
\DeclareRobustCommand\onedot{\futurelet\@let@token\@onedot}
\def\@onedot{\ifx\@let@token.\else.\null\fi\xspace}
\def\eg{\emph{e.g}\onedot}

\makeatother

\makeatletter
\def\bstctlcite{\@ifnextchar[{\@bstctlcite}{\@bstctlcite[@auxout]}}
\def\@bstctlcite[#1]#2{\@bsphack
  \@for\@citeb:=#2\do{%
    \edef\@citeb{\expandafter\@firstofone\@citeb}%
    \if@filesw\immediate\write\csname #1\endcsname{\string\citation{\@citeb}}\fi}%
  \@esphack}
\makeatother

\title{LLaCO: Continuous Diffusion for Large Language Models \\ with Competition-Aware Semantic Scheduling}

\renewcommand\affiliation[2][]{\addtolist[#1]{#2}{\affiliationlist}{\affiliationformat}{\par}}
\renewcommand{\affiliationfont}{\fontsize{9}{11}\selectfont}

\renewcommand{\authorlist}{%
  \authorformat[1*]{Zijian Jia},
  \authorformat[1*]{Peixuan Zhang},
  \authorformat[1]{Genying Li},
  \authorformat[1]{Di Wu},
  \authorformat[5]{Yujie Fu},\\[2pt]
  \authorformat[1,6\textnormal{\Letter}]{Si Li},
  \authorformat[3,4]{Boxin Shi},
  \authorformat[2,3\textnormal{\Letter}]{Shuchen Weng}%
}
\affiliation[1]{School of Artificial Intelligence, Beijing University of Posts and Telecommunications}
\affiliation[2]{Beijing Academy of Artificial Intelligence}
\affiliation[3]{State Key Laboratory of Multimedia Information Processing, Peking University}
\affiliation[4]{National Engineering Research Center of Visual Technology, Peking University}
\affiliation[5]{Yuanpei College, Peking University}
\affiliation[6]{Beijing Key Laboratory of Multimodal Data Intelligent Perception and Governance}

\contribution[*]{Equal contributions}
\contribution[\textnormal{\Letter}]{Corresponding authors}

\abstract{
Continuous diffusion models have emerged as a promising non-autoregressive paradigm for text generation, enabling bidirectional contextual planning and fine-grained semantic transitions over continuous embedding spaces. Along the continuous trajectory, the residual token identity in an intermediate state directly shapes the denoising objective at each step. However, the geometric trajectory strictly misaligns with temporal progression, triggering severe temporal compression near the noisy endpoint. 
In this paper, we show that embedding norm decay causes Euclidean distance to yield spurious overtaking, indicating that token competition must be evaluated directionally. In this space, a candidate overtakes the source token only when the effective noise magnitude surpasses their semantic separation. 
We formalize these pairwise dynamics as the \emph{Competition Threshold Theory} (CTT). Crucially, high-dimensional overtaking requires extreme noise levels, causing standard trajectories to concentrate identity decay into a narrow window near the noisy endpoint. To resolve this, we propose \emph{Semantic Retention Scheduling} (SRS) to align time allocation with the progression of identity decay, and demonstrate that existing token identity metrics are fundamentally statistical aggregations of these pairwise crossings. 
Building on these theoretical insights, we introduce \textbf{LLaCO}, a generative framework powered by competition-aware semantic scheduling. LLaCO adapts a pretrained backbone to continuous representations via an embedding directional adapter and timestep-conditioned modulation. To operate within the directional space, LLaCO employs a spherical noising process formulated as a geodesic interpolation toward a mask-biased endpoint, eliminating norm decay while preserving accurate token competition. Guided by SRS, LLaCO reallocates the denoising trajectory proportionally to the logarithmic rank decay, dedicating maximum optimization budget to critical identity transitions. 
Instantiated as an 8B model, LLaCO comprehensively outperforms previous continuous diffusion baselines across extensive benchmarks and rivals discrete diffusion models of comparable scale, laying a principled foundation for future exploration.
}

\MyEmail{\raggedright\mbox{\{jiazijian, pxzhang, lisi\}@bupt.edu.cn}, \mbox{\{shiboxin, shuchenweng\}@pku.edu.cn}\par}
\checkdata[Project Page]{\url{https://escapistmost.github.io/LLaCO/}}
\checkdata[Code]{\url{https://github.com/WhenGIX/LLaCO}}
\date{2026-10-22}

\begin{document}
\bstctlcite{IEEEexample:BSTcontrol}

\maketitle

\section{Introduction}
\label{sec:intro}

Beyond the dominant autoregressive~\cite{brown2020language,grattafiori2024llama3,yang2025qwen3} and discrete diffusion~\cite{sahoo2024simple,nie2025large,ye2025dream} paradigms, continuous diffusion models have emerged as a promising approach for generating text via iterative denoising~\cite{li2022diffusionlm,dieleman2022continuous,chen2026langflow}. They operate over continuous embedding spaces, enabling both bidirectional contextual planning and fine-grained semantic transitions~\cite{han2023ssdlm,shariatian2025ladd,zhou2025ccdd}. Along the continuous trajectory, the residual token identity in an intermediate state directly shapes the denoising objective at each step. Consequently, understanding how this identity decays along the trajectory is essential for designing principled corruption and denoising processes.

Prior works attempt to manage this identity decay. Some optimize time schedules based on loss or conditioning signals~\cite{dieleman2022continuous,chen2026langflow,yang2026replaid}, while others derive pacing from token identity metrics~\cite{pynadath2026candi,lee2026flowmap,deschenaux2026hyperspherical}.
Collectively, these studies reveal that identity decays as noise increases, where the source token is progressively outranked by competing candidates. They further observe that the geometric trajectory strictly misaligns with temporal progression, triggering severe temporal compression~\cite{lee2026flowmap,deschenaux2026hyperspherical}.
However, existing approaches primarily quantify the extent of empirical identity decay, lacking an analytical understanding of when and why competitors overtake the source.

In this paper, we show that embedding norm decay causes Euclidean distance to yield spurious overtaking, indicating that token competition must be evaluated directionally. In this space, a candidate overtakes the source token only when the effective noise magnitude surpasses their semantic separation. We formalize these pairwise dynamics as the \emph{Competition Threshold Theory} (Sec.~\ref{sec:CTT}).
Crucially, high-dimensional overtaking requires extreme noise levels. Standard trajectories thus trigger severe temporal compression, concentrating identity decay into a narrow window near the noisy endpoint. To resolve this, we reallocate timesteps along the diffusion trajectory to align with the progression of identity decay, denoted as \emph{Semantic Retention Scheduling} (Sec.~\ref{sec:SRS}).
Finally, we demonstrate that existing token identity metrics are fundamentally statistical aggregations of these pairwise crossings. Empirical measurements on pretrained embeddings closely track these theoretical derivations (Sec.~\ref{sec:val}).

Building on these theoretical insights, we introduce \textbf{LLaCO} (Continuous Diffusion for Large Language Models), a generative framework powered by \emph{Competition-Aware Semantic Scheduling}~(Sec.~\ref{sec:llaco}).
LLaCO adapts a pretrained backbone~\cite{fu2026nemotron} to continuous representations via an embedding directional adapter and timestep-conditioned modulation.
To operate within the directional space, LLaCO employs a spherical noising process. By formulating this process as a geodesic interpolation toward a mask-biased endpoint, we eliminate norm decay while preserving accurate token competition.
To overcome the aforementioned temporal compression, LLaCO reallocates the denoising trajectory proportionally to the logarithmic rank decay. This dedicates maximum optimization budget to identity transitions.
During inference, text is synthesized via sequential blockwise denoising along spherical geodesics, seamlessly translating continuous semantic trajectories into discrete tokens.

LLaCO establishes a principled continuous diffusion paradigm for large language models. We instantiate this framework as an 8B model. Across extensive benchmarks, it outperforms prior continuous diffusion baselines and rivals discrete diffusion models of comparable scale. We summarize our primary contributions below: 

\begin{itemize}

  \item We establish the Competition Threshold Theory to explain token competition, and propose Semantic Retention Scheduling to resolve the temporal compression problem.
  
  \item We introduce LLaCO based on the proposed principles, featuring an embedding directional adapter, timestep-conditioned modulation, spherical noising, and trajectory reallocation.
  
  \item LLaCO comprehensively outperforms previous continuous diffusion baselines and rivals discrete diffusion models, laying a principled foundation for future exploration.
\end{itemize}

\section{Related Work}
\label{sec:related_work}

\subsection{Autoregressive Language Models}

Large language models predominantly rely on the autoregressive Transformer architecture~\cite{vaswani2017attention,brown2020language,grattafiori2024llama3}, which models text generation as a sequence of next-token predictions under a causal attention mask~\cite{yang2025qwen3,deepseekai2026v4,touvron2023llama}.
While this causal formulation achieves leading generative capabilities, its inherently sequential inference couples decoding latency to sequence length~\cite{leviathan2023fast,gu2018non,stern2018blockwise}.
Furthermore, step-wise token commitments without native revision mechanisms can exacerbate compounding errors~\cite{ranzato2016sequence,wiseman2016sequence,arora2022exposure,dziri2023faith}.
Although acceleration strategies such as speculative decoding~\cite{leviathan2023fast,chen2023accelerating,li2026eagle} and multi-token prediction~\cite{gloeckle2024better,cai2024medusa,liu2024deepseek} alleviate inference latency, they remain inherently constrained by the underlying causal paradigm.

\subsection{Discrete Diffusion Language Models}
Discrete diffusion language models operate over discrete token spaces to govern corruption and generation dynamics~\cite{austin2021structured,lou2024discrete},  naturally enabling full-sequence planning and dynamic global revision.
To enhance capability and efficiency, recent efforts have scaled discrete diffusion by pretraining from scratch~\cite{nie2025large,nie2026improved} or adapting pretrained foundations~\cite{ye2025dream,gong2025scaling,bie2025llada2}. These advancements frequently incorporate blockwise decoding~\cite{arriola2025block,fu2026nemotron}, token re-masking~\cite{bie2026llada21}, and hybrid formulations~\cite{zheng2026cadd,zhou2025ccdd,shariatian2025ladd,pynadath2026candi}. 
Across these variations, the generation trajectory fundamentally relies on discrete state transitions, which inherently restricts continuous trajectory optimization and smooth gradient-based steering.

\subsection{Continuous Diffusion Language Models}
Continuous diffusion models generate text via iterative denoising directly within continuous embedding spaces, achieving competitive quality at small scales~\cite{chen2026langflow,yang2026replaid,hu2026elf,lee2026flowmap} and through pretrained adaptations~\cite{tae2025tess2,wang2026xprediction,yu2026dslllada}. 
A central challenge is the geometric misalignment between continuous trajectories and discrete vocabulary identities, which causes non-uniform identity decay~\cite{lee2026flowmap,deschenaux2026hyperspherical,jo2025rdlm}. 
To mitigate this degeneracy, prior approaches adjust noise schedules via empirical heuristics, such as loss monitoring or codebook analyses~\cite{dieleman2022continuous,chen2026langflow,pynadath2026candi,deschenaux2026hyperspherical,hu2026elf}, but lack an analytical understanding of how the underlying representation geometry governs this decay. 
To resolve this gap, we establish the Competition Threshold Theory and propose Semantic Retention Scheduling.

\makeatletter
\@ifundefined{jzjblock}{\newenvironment{jzjblock}{\par\color{blue}}{\par}}{}
\makeatother

\section{Theoretical Analysis} \label{sec:theoretical_analysis}

\subsection{Competition Threshold Theory} \label{sec:CTT}
\label{sec:token_competition}

\begin{figure}[t]
    \centering
    \includegraphics[width=\columnwidth]{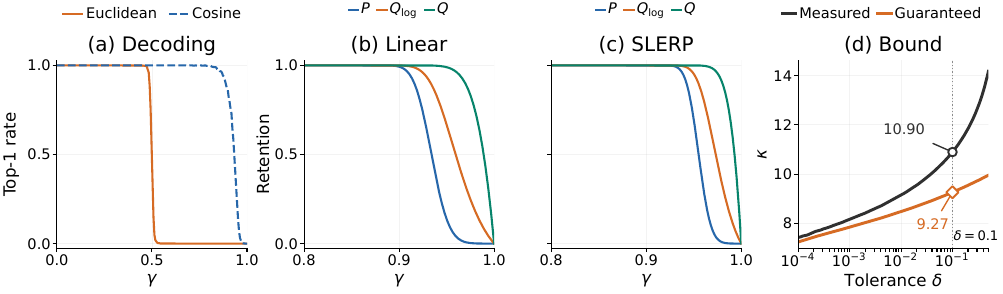}
    \caption{Identity decay dynamics along continuous noising trajectories.
    (a) Top-1 source-token retention across pretrained embeddings along a standard linear trajectory, comparing directional cosine ranking against Euclidean decoding. 
    (b) Endpoint-normalized retention statistics of unit token directions evaluated along the linear noising trajectory. 
    (c) Endpoint-normalized retention statistics of unit token directions evaluated along the spherical geodesic trajectory. 
    (d) Comparison between the preservation scale guaranteed by our full-spectrum theoretical bound and the empirically measured location where identity decay reaches tolerance $\delta$, evaluated across varying tolerance thresholds $\delta$ with markers highlighting $\delta = 0.1$.
    See Appendices~\ref{app:competition_time_comparison} and \ref{app:empirical_preservation_bounds} for details.}
\label{fig:identity_time}
\end{figure}

\noindent \textbf{Directional identity.}
\label{sec:directional_identity}
The progression of identity decay is fundamentally governed by pairwise competition against other candidates (Appendix~\ref{app:information_competition}). 
As shown in Fig.~\ref{fig:identity_time}~(a), conventional Euclidean decoding causes the source token to yield premature spurious overtaking compared to cosine ranking. 
To understand this discrepancy, consider a corrupted state $s = \alpha e_i + \beta z$ for a source token $i$ with direction $u_i = e_i / \|e_i\|$, where $\alpha > 0$ denotes signal scale, $\beta \ge 0$ noise scale, and $z$ a fixed noise realization. 
Decoding by minimal Euclidean distance $\|s - e_j\|^2$ is equivalent to ranking candidates by the following decoupled score:
\begin{equation}
    \operatorname{Score}_{\mathrm{Euc}}(j) = 2 n_j u_j^\top s - n_j^2,
\end{equation}
where $n_j = \|e_j\| > 0$ and $u_j = e_j / n_j$. 
Crucially, this score decouples into an \emph{alignment term} ($2 n_j u_j^\top s$) that shrinks with the signal $\alpha$, and a \emph{norm penalty} ($-n_j^2$) that remains entirely static. 
This asymmetry leads to a severe degeneracy. Even under zero noise ($\beta = 0$), where the state $s = \alpha e_i$ points perfectly along the true source direction $u_i$, any candidate with a smaller norm ($n_j < n_i$) will inevitably overtake the source as $\alpha \to 0$ (Appendix~\ref{app:euclidean_pathology}).
Specifically, at $\beta = 0$ the score gap $\operatorname{Score}_{\mathrm{Euc}}(i) - \operatorname{Score}_{\mathrm{Euc}}(j) = 2\alpha n_i (n_i - n_j u_i^\top u_j) - (n_i^2 - n_j^2)$ is affine in $\alpha$ and changes sign at $\alpha^0_{ij} = (n_i^2 - n_j^2) / [2 n_i (n_i - n_j u_i^\top u_j)] \in (0, 1)$, making the candidate overtake the source once $\alpha < \alpha^0_{ij}$.
This spurious overtaking is entirely driven by embedding norm decay rather than true noise. To eliminate this effect, we rank by the cosine score $c_j=u_j^\top s/\|s\|$, ensuring that only directional noise drives competition.

\noindent \textbf{Competition thresholds.}
To formalize when a candidate overtakes the source token, define the semantic similarity $a_{ij} = u_i^\top u_j$, the noise projection $b_j = u_j^\top z$, and the effective noise magnitude $\kappa = \beta / (\alpha n_i)$. For any candidate distinct from the source ($a_{ij} < 1$), the cosine score gap expands as:
\begin{equation}
    c_j - c_i = \frac{\alpha n_i}{\|s\|} \Big[ -(1 - a_{ij}) + \kappa (b_j - b_i) \Big],
    \label{eq:competition_gap}
\end{equation}
where $1 - a_{ij} > 0$ represents the intrinsic semantic separation, and $b_j - b_i$ captures the relative noise advantage.
Candidate $j$ overtakes the source precisely when this score gap becomes positive ($c_j > c_i$). Since this occurs if and only if the noise drives $b_j > b_i$ and $\kappa$ exceeds a critical threshold, we define the pairwise competition threshold as:
\begin{equation}
    \tau_{ij} = 
    \begin{cases}
        \dfrac{1 - a_{ij}}{b_j - b_i}, & \text{if } b_j > b_i, \\[6pt]
        +\infty, & \text{otherwise}.
    \end{cases}
    \label{eq:pairwise_threshold}
\end{equation}
Aggregating pairwise threshold-crossing events determines the instantaneous rank of the source token $r_i(\kappa)$ and its first-overtaking threshold $\kappa_i^*$:
\begin{equation}
    r_i(\kappa) = 1 + \sum_{j \neq i} \mathbb{I}\{\tau_{ij} < \kappa\}, \quad \kappa_i^* = \min_{j \neq i} \tau_{ij}.
    \label{eq:rank_after_overtaking}
\end{equation}
\begin{proposition}[Competition thresholds]
\label{prop:competition_thresholds}
Fix a source token $i$ and noise vector $z$ with signal scale $\alpha > 0$. A candidate $j$ overtakes the source if and only if $\kappa > \tau_{ij}$. Consequently, the rank $r_i(\kappa)$ is monotonically non-decreasing in $\kappa$, and the source token remains top-1 if and only if $\kappa \le \kappa_i^*$ (Appendix~\ref{app:threshold_details}).
\end{proposition}

\noindent \textbf{Pairwise overtaking.}
Along any continuous diffusion trajectory, the noise $z$ is stochastic, making each competition threshold $\tau_{ij}$ an inherent random variable. 
For a random source $i$ and noise distribution $z \sim \mathcal{D}(\cdot \mid i)$, the linearity of expectation decomposes the global expected rank into pairwise overtaking probabilities:
\begin{equation}
    \mathbb{E}[r(\kappa)] = 1 + \mathbb{E}_i \left[ \sum_{j \neq i} \Pr_z[\tau_{ij} < \kappa \mid i] \right].
    \label{eq:expected_rank}
\end{equation}
Under standard noise $z \sim \mathcal{N}(0, I_d / d)$, the relative noise advantage follows $b_j - b_i \sim \mathcal{N}(0, 2(1 - a_{ij}) / d)$. Evaluating the probability condition $\tau_{ij} < \kappa$ yields the closed-form overtaking probability:
\begin{equation}
    p_{ij}(\kappa) = \bar{\Phi}\left( \frac{1}{\kappa}\sqrt{\frac{d(1 - a_{ij})}{2}} \right),
    \label{eq:pairwise_prob}
\end{equation}
where $\bar{\Phi}$ is the complementary CDF of the standard Gaussian distribution (Appendix~\ref{app:gaussian_competition}).
Together, Eqs.~\eqref{eq:expected_rank} and~\eqref{eq:pairwise_prob} show that expected rank decay is governed entirely by the embedding similarity spectrum $\{a_{ij}\}$ and effective noise magnitude $\kappa$, with the overtaking probability $p_{ij}$ monotonically increasing in both $a_{ij}$ and $\kappa$ (Fig.~\ref{fig:prediction_measurement}~(a)).

\noindent \textbf{Preservation guarantee.}
Aggregating these pairwise overtaking probabilities through a union bound yields a sufficient condition under which the source token remains top-1 with high probability (Appendices~\ref{app:preservation_shared} and~\ref{app:empirical_preservation_bounds}).
Let $P_{\mathrm{raw}}(\kappa \mid i) = \Pr_z[\kappa_i^* \ge \kappa \mid i]$ denote the raw top-1 retention of source token $i$. 
Because identity decay occurs once any candidate overtakes the source, the probability of decay is bounded by the sum of pairwise overtaking probabilities:
\begin{equation}
    1 - P_{\mathrm{raw}}(\kappa \mid i) \le \sum_{j \neq i} p_{ij}(\kappa).
    \label{eq:main_union_bound}
\end{equation}
Substituting Eq.~\eqref{eq:pairwise_prob} into Eq.~\eqref{eq:main_union_bound} and bounding pairwise similarities by the maximal candidate similarity $a_{\max} = \max_{j \neq i} a_{ij} < 1$, we enforce $1 - P_{\mathrm{raw}}(\kappa \mid i) \le (V - 1)\,\bar{\Phi}\big(\frac{1}{\kappa}\sqrt{d(1 - a_{\max})/2}\big) \le \delta$ for any failure tolerance $\delta \in (0, 1)$. 
Inverting this condition yields a closed-form preservation scale:
\begin{equation}
    \kappa_\delta^{\mathrm{bd}} = \frac{\sqrt{d(1 - a_{\max})/2}}{\bar{\Phi}^{-1}\big(\delta / (V - 1)\big)},
    \label{eq:main_preservation_bound}
\end{equation}
such that any noise scale satisfying $\kappa \le \kappa_\delta^{\mathrm{bd}}$ rigorously guarantees $P_{\mathrm{raw}}(\kappa \mid i) \ge 1 - \delta$. 
Using the standard Gaussian tail asymptotic $\bar{\Phi}^{-1}(q) \sim \sqrt{2 \ln(1/q)}$ as $q \to 0$, this critical scale exhibits the scaling behavior:
\begin{equation}
    \kappa_\delta^{\mathrm{bd}} \approx \sqrt{\frac{d(1 - a_{\max})}{4 \ln((V - 1)/\delta)}},
\end{equation}
which grows as $\sqrt{d}$ while depending only logarithmically on the vocabulary size $V$. 
This theoretical scaling formally proves that high-dimensional overtaking requires extreme noise levels, thereby inducing temporal compression near the noisy endpoint. 
While this analytical bound provides a provable safety margin, its union-bound nature is conservative (Fig.~\ref{fig:identity_time}~(d)), motivating the exact empirical aggregate metrics detailed in Sec.~\ref{sec:SRS}.

\subsection{Semantic Retention Scheduling} 
\label{sec:SRS}

Standard diffusion trajectories severely compress this $\kappa$-governed competition into a narrow terminal time window. Therefore, we construct a principled scheduling framework that aligns generative time allocation with the progression of identity decay.

\noindent \textbf{Retention statistics.}
\label{sec:rank_statistics}
A source token's identity does not vanish instantaneously upon losing its top-1 rank; its relative rank continues to decline throughout diffusion.
Since rank simply counts the total number of candidates that have overtaken the source, different retention metrics correspond to different weighting schemes across these overtaking events.
Consider a trajectory parameterized by $\gamma \in [0, 1]$ toward the noisy endpoint with non-decreasing effective noise magnitude $\kappa(\gamma)$ and initial rank $r(0) = 1$.
Let $M(\gamma) = r(\gamma) - 1$ denote the number of overtaking events at step $\gamma$, increasing monotonically from $0$ to a terminal count $M_{\mathrm{end}} = r(1) - 1$.
On valid trajectories where identity decay occurs ($E = \{M_{\mathrm{end}} > 0\}$), any normalized retention metric must satisfy boundary conditions $R(0) = 1$ and $R(1) = 0$.
Weighting the cumulative overtaking count and normalizing by its terminal value naturally satisfies these boundary conditions, yielding the general form $R(\gamma) = 1 - f(M(\gamma)) / f(M_{\mathrm{end}})$ for any non-decreasing weighting function $f$ with $f(0) = 0$ and $f(M_{\mathrm{end}}) > 0$.
Instantiating $f$ with the indicator, identity, and logarithmic kernels yields top-1 retention $P(\gamma)$, linear retention $Q(\gamma)$, and our proposed logarithmic retention $Q_{\log}(\gamma)$:
\begin{equation}
\begin{split}
    P(\gamma) &= \mathbb{E}\big[\mathbb{I}\{M(\gamma) = 0\} \mid E\big], \\
    Q(\gamma) &= \mathbb{E}\left[1 - \frac{M(\gamma)}{M_{\mathrm{end}}} \,\Big|\, E\right], \\
    Q_{\log}(\gamma) &= \mathbb{E}\left[1 - \frac{\ln(1 + M(\gamma) / 2)}{\ln(1 + M_{\mathrm{end}} / 2)} \,\Big|\, E\right],
\end{split}
\label{eq:trajectory_retention}
\end{equation}
where $q(\gamma)$ and $q_{\log}(\gamma)$ denote the per-trajectory realizations inside the expectations.
The normalized metric $P(\gamma)$ relates to the source-averaged raw retention $P_{\mathrm{raw}}(\gamma)$ via $P = (P_{\mathrm{raw}} - P_{\mathrm{raw}}(1)) / (1 - P_{\mathrm{raw}}(1))$. Since the terminal retention $P_{\mathrm{raw}}(1) \approx 1/V$ is negligible for large vocabularies, $P$ and $P_{\mathrm{raw}}$ differ by at most $\mathcal{O}(1/V)$ and are practically interchangeable in subsequent derivations.

\begin{proposition}[Ordering of retention statistics]
\label{prop:ordering_retention}
Along any trajectory with non-decreasing $\kappa(\gamma)$ and $M_{\mathrm{end}} > 0$, the per-trajectory retentions satisfy $\mathbb{I}\{M(\gamma) = 0\} \le q_{\log}(\gamma) \le q(\gamma)$, which implies:
\begin{equation}
    P(\gamma) \le Q_{\log}(\gamma) \le Q(\gamma).
    \label{eq:rank_statistic_order}
\end{equation}
Furthermore, for any retention fraction $\eta \in (0, 1)$, $q_{\log}(\gamma) \le \eta$ if and only if $M(\gamma) \ge 2 \big[ (1 + M_{\mathrm{end}} / 2)^{1 - \eta} - 1 \big]$, whereas $q(\gamma) \le \eta$ if and only if $M(\gamma) \ge (1 - \eta) M_{\mathrm{end}}$.
\end{proposition}

The upper bound in Eq.~\eqref{eq:rank_statistic_order} follows from the concavity of the logarithm, while the lower bound holds because $q_{\log}$ remains at unity before the very first overtaking event (Appendix~\ref{app:rank_normalization}). 
Crucially, when the terminal overtaking count $M_{\mathrm{end}}$ is large (\eg, $\mathbb{E}[M_{\mathrm{end}}] = (V - 1)/2$ under isotropic endpoints), reaching half retention ($\eta = 0.5$) under the logarithmic metric requires only $M(\gamma) \approx \sqrt{2 M_{\mathrm{end}}}$ overtaking events, in contrast to the substantial $M_{\mathrm{end}} / 2$ required under linear retention.
Thus, while top-1 retention $P$ collapses entirely at the first overtake and linear retention $Q$ is diluted by the vast vocabulary tail, $Q_{\log}$ precisely captures the critical intermediate decay of leading semantic ranks.

\begin{figure}[t]
    \centering
    \includegraphics[width=\columnwidth]{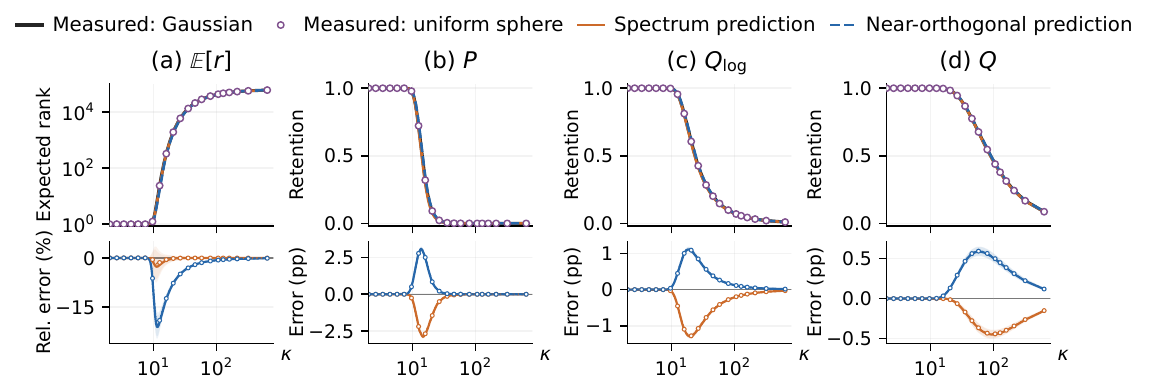}
      \caption{Predicted and measured competition statistics on pretrained embeddings against the effective noise magnitude $\kappa$: (a) expected rank $\mathbb{E}[r]$, (b) top-1 retention $P$, (c) logarithmic retention $Q_{\log}$, and (d) linear retention $Q$. Retention statistics are endpoint-normalized. Lower panels show prediction errors; shading denotes 95\% bootstrap intervals.}

    \label{fig:prediction_measurement}
    \vspace{-2mm}
\end{figure}

\noindent \textbf{Temporal compression.}
\label{sec:identity_window}
Pretrained token embeddings exhibit a heterogeneous high-dimensional geometry. They consist of a near-orthogonal bulk with only sparse localized semantic clusters~\cite{cai2013angles,guha2026representational} (Appendix~\ref{app:multimodel_geometry}).
This geometry persists across eight pretrained language models spanning autoregressive and diffusion paradigms, where over 94\% of token pairs are nearly orthogonal ($|a_{ij}| \le 0.1$) and each source token faces only a sparse set of semantically close competitors (Table~\ref{tab:multimodel_geometry}).
Under this near-orthogonal surrogate with effective vocabulary size $N = V - 1$, the extreme value concentration of the residual noise maximum $m_N = \mathbb{E}[\max_{j \neq i} Z_j] \sim \sqrt{2\ln N}$ over i.i.d. standard Gaussian residuals $Z_j$ yields (Appendix~\ref{app:first_overtaking}):
\begin{equation}
\begin{split}
    P(\kappa) &\approx \Phi\left(\frac{\sqrt{d}}{\kappa} - m_N\right), \\
    \kappa_{0.5} &\approx \frac{\sqrt{d}}{m_N} \sim \sqrt{\frac{d}{2\ln N}}.
\end{split}
\label{eq:kappa_half}
\end{equation}
This reveals a fundamental geometric scale. Because modern language models operate in the high-dimensional regime $d \gg \ln N$ (\eg, $d \approx 4096 \gg \ln N \approx 10$), initial identity decay cannot occur at moderate noise levels. 
Instead, overtaking requires an extraordinarily large effective noise magnitude $\kappa_{0.5} \gg 1$ (top-1 retention in Fig.~\ref{fig:prediction_measurement}~(b)).
Within this surrogate, the approximation error of $P(\kappa)$ in Eq.~\eqref{eq:kappa_half} remains strictly below $0.009$ for $d = 4096$ and $N = 130{,}067$. 
Evaluating $m_N$ numerically at this finite $N$ yields $m_N \approx 4.44$ (below its asymptotic value $\sqrt{2\ln N} \approx 4.85$), leading to $\kappa_{0.5} \approx 14.4$, which closely tracks the empirical measurement of $14.2$ on pretrained embeddings (Appendix~\ref{app:first_overtaking}).
Because the first-overtaking threshold $\kappa_i^*$ has an infinite mean even when conditioned on being finite (Appendix~\ref{app:first_overtaking}), we characterize this transition by quantiles, namely its half-retention location $\kappa_{0.5}$ and its transition width:
\begin{equation}
    W_\kappa = \kappa_{0.1} - \kappa_{0.9} \approx \frac{2 z_{0.9} \sqrt{d}}{m_N^2 - z_{0.9}^2},
    \label{eq:main_kappa_width}
\end{equation}
where $z_{0.9} = \Phi^{-1}(0.9)$. Both the location $\kappa_{0.5}$ and width $W_\kappa$ scale as $\sqrt{d}$, with the width governed primarily by fluctuations of the noise projection onto the source direction rather than by the residual noise maximum, which tightly concentrates around $m_N$ (Appendix~\ref{app:first_overtaking}).
To understand how this spatial requirement manifests over the generative trajectory, we formally decouple spatial decay from temporal scheduling.

\begin{corollary}[Trajectory factorization]
\label{cor:path_factorization}
Recall from Proposition~\ref{prop:competition_thresholds} that pairwise thresholds $\tau_{ij}$ depend solely on embedding geometry and noise realizations, entirely independent of the time coordinate $\gamma$. 
Moreover, since $\kappa(1) = \infty$ at the noisy endpoint, the terminal count $M_{\mathrm{end}} = |\{j : b_j > b_i\}|$ and the valid event $E$ are likewise independent of the trajectory.
Consequently, under any sample-independent effective-noise schedule $\kappa(\gamma)$, any macroscopic retention metric $R \in \{P, Q, Q_{\log}\}$ factorizes as:
\begin{equation}
    R(\gamma) = \mathcal{R}_{\mathcal{D}}\big(\kappa(\gamma)\big),
    \label{eq:path_factorization}
\end{equation}
where the outer function $\mathcal{R}_{\mathcal{D}}(\kappa)$ denotes the intrinsic spatial retention governed strictly by representation geometry and noise distribution, and the inner scalar function $\kappa(\gamma)$ serves as a coordinate ruler mapping spatial noise scales onto the trajectory interval $[0, 1]$.
\end{corollary}
This factorization immediately exposes the structural degeneracy of standard diffusion trajectories. 
For a standard linear trajectory $s_\gamma = (1 - \gamma) u_i + \gamma z$, the effective noise magnitude is a highly convex function $\kappa(\gamma) = \gamma / (1 - \gamma)$, which inverts to the remaining trajectory fraction $1 - \gamma = 1 / (1 + \kappa)$ (Fig.~\ref{fig:identity_time}~(b)). 
Because the high-dimensional geometry $d \gg \ln N$ demands an exceptionally large $\kappa_{0.5}$, this convex coordinate mapping forces the top-1 rank decay and its $\mathcal{O}(d^{-1/2})$-wide transition window to collapse into the final $\mathcal{O}(\sqrt{\ln N / d})$ residual fraction of the trajectory (Appendix~\ref{app:path_factorization}). 
At a fixed vocabulary size, increasing the dimension $d$ widens the transition along the effective noise magnitude, yet narrows it in trajectory time and shifts it toward the noisy endpoint.
As a result, uniform geometric sampling allocates over $90\%$ of its optimization budget to early, stagnant retention phases while severely undersampling the narrow terminal window where semantic identities actually transition.


\subsection{Empirical Validation} 
\label{sec:val}

\noindent \textbf{Theoretical predictions via noise decoupling.}
While pairwise probabilities (Eq.~\eqref{eq:pairwise_prob}) capture marginal competition, macroscopic retention depends on joint candidate behavior. 
Under isotropic Gaussian noise, we decouple this joint competition by decomposing the noise vector into a shared component along the source direction and orthogonal residuals (Appendix~\ref{app:source_projection}).

Let $\sigma_b = d^{-1/2}$, $Y = z^\top u_i / \sigma_b \sim \mathcal{N}(0, 1)$ denote the normalized shared source projection, $h = 1 / (\kappa \sigma_b)$, and $\rho_{ij} = \sqrt{(1 - a_{ij}) / (1 + a_{ij})}$. 
Decomposing each candidate into source-parallel and orthogonal components reveals that candidate $j$ overtakes the source if and only if $Z_j > \rho_{ij}(Y + h)$, where $Z_j \sim \mathcal{N}(0, 1)$ is an independent standard Gaussian residual ($Z_j \perp Y$) and the shared projection $Y$ simultaneously couples all competing candidates (Appendix~\ref{app:source_projection}).
Conditioned on a realized shared projection $Y = y$, linearity of expectation yields the exact conditional expectation for the number of overtakes: 
\begin{equation}
    \mu_i(y, \kappa) = \sum_{j \neq i} \bar{\Phi}\big(\rho_{ij}(y + h)\big).
    \label{eq:conditional_count}
\end{equation}
Treating the residuals as mutually independent and integrating over the Gaussian density $\phi(y)$ of $Y$ yields spectrum-aware analytical predictions for all three retention statistics:
\begin{equation}
\begin{split}
        P_{\mathrm{raw}}(\kappa|i) &\approx \!\int\! \phi(y) \prod_{j\ne i} \Phi(\rho_{ij}(y+h))\, \mathrm{d}y, \\
    Q(\kappa) &\approx \mathbb{E}\left[ 1 - \frac{\mu}{\mu^T} \,\Big|\, E \right], \\
    Q_{\log}(\kappa) &\approx \mathbb{E}\left[ 1 - \frac{\ln(1 + \mu/2)}{\ln(1 + \mu^T/2)} \,\Big|\, E \right],
\end{split}
\label{eq:spectrum_predictions}
\end{equation}
where $\mu = \mu_i(Y, \kappa)$ and $\mu^T = \mu_i(Y, \infty)$ denote the conditional means of $M(\gamma)$ and $M_{\mathrm{end}}$, replacing the random counts within the ratios of $Q$ and $Q_{\log}$. 
Conditioning on $E$ weights the integral over $Y$ by $\Pr(E \mid i, Y) = 1 - \prod_{j \neq i} \Phi(\rho_{ij} Y)$, with full derivations provided in Appendix~\ref{app:qlog_conditional}. 
The endpoint-normalized metric $P$ directly follows via $P = [P_{\mathrm{raw}} - P_{\mathrm{raw}}(1)] / [1 - P_{\mathrm{raw}}(1)]$. 
These predictions depend solely on $\kappa$ and the similarity spectrum $\{a_{ij}\}$, closely matching empirical data across all three statistics as shown in Fig.~\ref{fig:prediction_measurement}~(b-d). Setting $\rho_{ij} = 1$ further yields closed-form near-orthogonal asymptotic surrogates (Appendix~\ref{app:near_orthogonal}):
\begin{equation}
\begin{split}
    \mathbb{E}[r^\perp(\kappa)] &= 1 + N \bar{\Phi}(h/\sqrt{2}), \\
    P_{\mathrm{raw}}^\perp(\kappa) &= \!\int\! \phi(y)\, \Phi(y+h)^N \, \mathrm{d}y, \\
    Q^\perp(\kappa) &= \!\int\! w_N(y) \left[ 1 - \frac{p(y,\kappa)}{p_T(y)} \right] \mathrm{d}y, \\
    Q_{\log}^\perp(\kappa) &\approx \!\int\! w_N(y) \left[ 1 - \frac{\ln(1 + N p(y,\kappa)/2)}{\ln(1 + N p_T(y)/2)} \right] \mathrm{d}y,
\end{split}
\label{eq:main_near_orthogonal_predictions}
\end{equation}
where $p(y,\kappa) = \bar{\Phi}(y+h)$ and $p_T(y) = \bar{\Phi}(y)$ are the per-candidate overtaking probabilities at $\kappa$ and at the endpoint, and $w_N(y) = \frac{N+1}{N}\phi(y)\big[1 - \Phi(y)^N\big]$ is the density of $Y$ conditioned on $E$. The endpoint-normalized retention is $P^\perp = [(N+1)P_{\mathrm{raw}}^\perp - 1]/N$. At any given noise magnitude $\kappa$, these closed-form surrogates depend solely on the fundamental architectural dimensions $d$ and $V$.

\noindent \textbf{Empirical validation on pretrained embeddings.}
We validate these theoretical predictions on the pretrained token embeddings~\cite{fu2026nemotron}, measuring retention statistics across both standard Gaussian noise endpoints and uniform spherical endpoints (Appendix~\ref{app:embedding_validation}).
Remarkably, without parameter fitting, both predictions track empirical measurements with exceptional precision. The half-retention errors remain strictly within 2\% (Fig.~\ref{fig:prediction_measurement}; Appendix~\ref{app:competition_comparison}).
Retaining the full similarity spectrum $\{a_{ij}\}$ further reduces the early-phase discrepancies caused by a handful of high-similarity semantic neighbors.
In contrast, the preservation guarantee in Eq.~\eqref{eq:main_union_bound} provides a provably conservative safety margin, ending strictly before the observed transition (Fig.~\ref{fig:identity_time}~(d); Appendix~\ref{app:empirical_preservation_bounds}).
Specifically, for tolerance $\delta = 0.1$, the full-spectrum bound certifies $\kappa \le 9.27$ (where empirical identity decay is merely 1.24\%), whereas the measured decay reaches 10\% only at $\kappa = 10.90$.
Empirical retention curves for Gaussian and uniform spherical endpoints nearly coincide when plotted against the effective noise magnitude $\kappa$.
Theoretically, projecting a Gaussian endpoint $z_G \sim \mathcal{N}(0, I_d/d)$ onto the unit sphere rescales competition thresholds samplewise as $\tau_{ij}^{S} = \|z_G\|\,\tau_{ij}^{G}$; because $\|z_G\|$ tightly concentrates around $1$ in high dimensions ($d=4096$), the two retention profiles naturally coincide (Appendix~\ref{app:gaussian_competition}).
This supports the effective-noise characterization of token competition under isotropic endpoints. The observed divergence between linear and spherical trajectories over trajectory time $\gamma$ arises primarily from their distinct temporal mappings $\kappa(\gamma)$ (Fig.~\ref{fig:identity_time}(b, c); Appendix~\ref{app:competition_time_comparison}).

\section{LLaCO}
\label{sec:llaco}

\begin{figure}[t]
    \centering
    \includegraphics[width=\columnwidth]{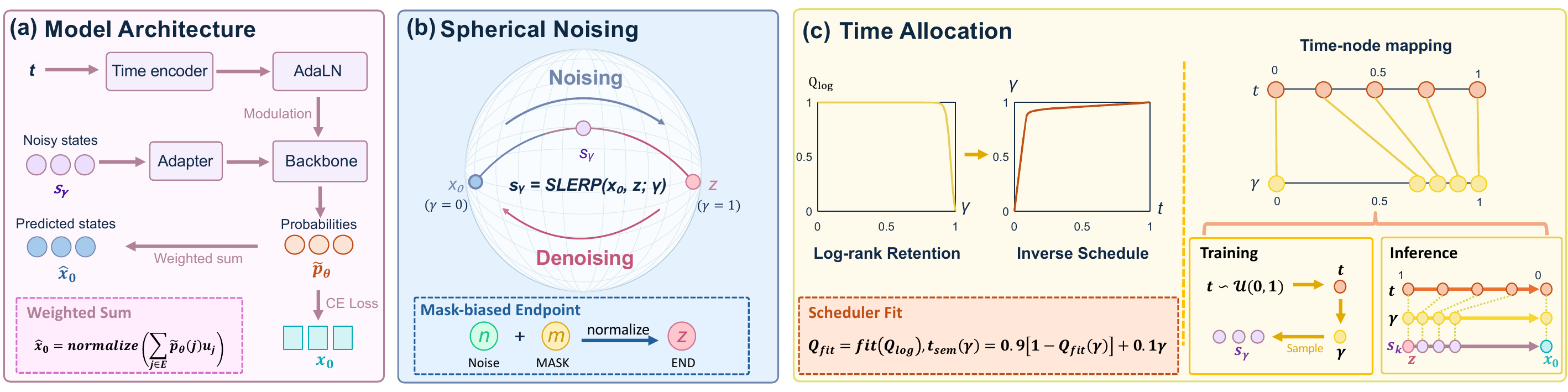}

    \caption{
    Overview of the LLaCO framework. (a) The model architecture features an embedding directional adapter and timestep-conditioned modulation to produce token probabilities for cross-entropy training and clean-direction estimation. (b) Spherical noising proceeds along a geodesic trajectory toward a mask-biased endpoint. (c) Semantic Retention Scheduling leverages logarithmic rank retention to reallocate the optimization budget during both training and inference. Here, $t$ denotes $t_{\mathrm{sem}}$ for simplicity.
    }
    \label{fig:llaco_pipeline}
    \vspace{-2mm}
\end{figure}

\subsection{Model Architecture}
\label{sec:model_architecture}

To adapt a pretrained transformer backbone for continuous directional states, LLaCO introduces an embedding directional adapter and timestep-conditioned modulation, optimized via multi-block cross-entropy (Fig.~\ref{fig:llaco_pipeline}(a); Appendix~\ref{app:denoising_implementation}).

\noindent \textbf{Directional adapter.}
Directly feeding continuous states into the backbone induces a severe distribution shift from the discrete pretrained embeddings.
We deploy an embedding directional adapter that projects a unit continuous state $s$ back onto the backbone's semantic manifold via a soft convex combination of original embeddings, $\widetilde{e}(s) = \sum_{j=1}^V w_j(s) e_j$, with normalized weights:
\begin{equation}
    w_j(s) = \operatorname{softmax}_j\left(\eta s^\top u_j + h_j\right),
    \label{eq:directional_adapter}
\end{equation}
where $u_j = e_j / \|e_j\|_2$ denotes the normalized direction of token $j$, and $\eta \in \mathbb{R}^+$ and $h_j \in \mathbb{R}$ are learnable inverse-temperature and bias parameters.

\noindent \textbf{Timestep-conditioned modulation.}
To inform the network of the corruption stage, the hidden states are conditioned on a semantic time coordinate $t_{\mathrm{sem}} \in [0, 1]$, mapped from the trajectory coordinate $\gamma$ via Semantic Retention Scheduling (Sec.~\ref{sec:SRS}).
A sinusoidal time encoder with an MLP maps $t_{\mathrm{sem}}$ to a conditioning vector $c$. At transformer layer $\ell$, zero-initialized linear projections generate adaptive shift $\delta$, scale $\sigma$, and gate $g$ modulations for each sublayer $F$:
\begin{equation}
    y = x + (1 + g) \odot F\big( \operatorname{Norm}(x) \odot (1 + \sigma) + \delta \big),
    \label{eq:time_modulation}
\end{equation}
where $[\delta, \sigma, g] = W_\ell^F \operatorname{LN}(c)$, $\operatorname{LN}(\cdot)$ denotes layer normalization, $\operatorname{Norm}(\cdot)$ is the backbone's pre-sublayer normalization, and $\odot$ is element-wise multiplication. Zero-initialization preserves the pretrained backbone's sublayer transformations at initialization.

\noindent \textbf{Training objective.}
Token embeddings remain frozen, while the adapter, timestep-conditioned modulation, and backbone are trained end-to-end. For target block $y_b$ with continuous state $s_\gamma^{(b)}$ at semantic time $t_{\mathrm{sem}}^{(b)}$ and preceding clean context $x_{<b}$ (encompassing prompt context and previous clean blocks), the model minimizes the cross-entropy loss:
\begin{equation}
    \mathcal{L} = -\mathbb{E}\left[ \log p_\theta\big(y_b \mid x_{<b}, \widetilde{e}(s_\gamma^{(b)}), t_{\mathrm{sem}}^{(b)}\big) \right],
    \label{eq:denoising_loss}
\end{equation}
where $y_b$ denotes the ground-truth token sequence in block $b$, and the expectation is taken over data batches, blocks, and sampled times. Detailed blockwise attention masking and parallel training implementation appear in Appendix~\ref{app:denoising_implementation}.


\subsection{Spherical Noising}
\label{sec:spherical_noising}

As established in the Competition Threshold Theory (Sec.~\ref{sec:CTT}), the effective noise magnitude $\kappa = \beta / (\alpha n_i)$ is confounded by token norm variations under standard linear trajectories, while embedding norm decay induces spurious overtaking.
To eliminate these artifacts,is confounded by token norm variations under standard linear trajectories, causing Euclidean norm decay to induce spurious overtaking.
To eliminate this artifact, LLaCO corrupts unit representations on the unit hypersphere $\mathbb{S}^{d-1}$ (Fig.~\ref{fig:llaco_pipeline}~(b)).

\noindent \textbf{Spherical geodesic.}
Given a clean unit direction $u_i$ and a unit endpoint $z \in \mathbb{S}^{d-1}$ with angle $\theta = \arccos(u_i^\top z)$, the intermediate state evolves along the spherical geodesic (SLERP)~\cite{shoemake1985animating}:
\begin{equation}
    s_\gamma = \frac{\sin((1-\gamma)\theta)}{\sin\theta} u_i + \frac{\sin(\gamma\theta)}{\sin\theta} z,
    \label{eq:spherical_path}
\end{equation}
where $\gamma\in[0,1]$ is the fraction of the arc traversed. Along this trajectory, the instantaneous effective noise magnitude satisfies:
\begin{equation}
    \kappa(\gamma) = \frac{\sin(\gamma\theta)}{\sin((1-\gamma)\theta)},
    \label{eq:spherical_noise_ratio}
\end{equation}
with derivative $\kappa'(\gamma) = \theta\sin\theta / \sin^2((1-\gamma)\theta) > 0$ for non-antipodal endpoints ($0 < \theta < \pi$). Thus, $\kappa(\gamma)$ strictly increases and diverges as $\gamma \to 1$, ensuring pathwise rank monotonicity (Appendices~\ref{app:path_factorization} and~\ref{app:mask_endpoint}).

\noindent \textbf{Mask-biased endpoint distribution.}
Rather than drawing purely isotropic endpoints, we incorporate inductive bias from the pretrained discrete language model by sampling the terminal point as:
\begin{equation}
    z = \operatorname{normalize}(n + \lambda m), \qquad \lambda = 1,
    \label{eq:mask_endpoint}
\end{equation}
where $n \sim \operatorname{Unif}(\mathbb{S}^{d-1})$ provides isotropic stochasticity, and $m$ is the unit embedding of the \texttt{[MASK]} token. In high dimensions, $n^\top m \approx 0$ with high probability, so $z$ forms an angle of $\approx 45^\circ$ with $m$. 
Accordingly, $z \approx (n + \lambda m)/\sqrt{1 + \lambda^2}$: the isotropic component is scaled by $1/\sqrt{1 + \lambda^2}$, while the deterministic component $\lambda m/\sqrt{1 + \lambda^2}$ persists after averaging over $n$.
This mask bias shifts the mean relative noise advantage $b_j - b_i$ defined in Eq.~\eqref{eq:competition_gap} by:
\begin{equation}
    \Delta\mu_j = \frac{\lambda}{\sqrt{1+\lambda^2}}\big(m^\top u_j - m^\top u_i\big).
    \label{eq:mask_mean_shift}
\end{equation}
Since the relative noise advantage remains approximately Gaussian, token competition under this endpoint retains the pairwise form of Eq.~\eqref{eq:pairwise_prob}. For $\lambda = 1$, the pairwise overtaking probability becomes
\begin{equation}
    p_{ij}(\kappa) \approx \bar{\Phi}\left(\frac{(1 - a_{ij})/\kappa - \Delta\mu_j}{s_{ij}}\right),
    \label{eq:main_mask_pairwise}
\end{equation}
where $s_{ij}^2 = (1 - a_{ij})/d - 3\big(m^\top(u_j - u_i)\big)^2/(8d)$ denotes the variance of the relative noise advantage, which normalization contracts to at most half of its isotropic counterpart.
Candidates with higher semantic affinity to the mask direction are more likely to overtake the source, while isotropic randomness ensures broad coverage (Appendix~\ref{app:mask_endpoint}). Source tokens and ranks range over the valid vocabulary, excluding \texttt{[MASK]}.

\subsection{Trajectory Reallocation}
\label{sec:semantic_time}

Guided by Semantic Retention Scheduling (Sec.~\ref{sec:SRS}), we decouple the geometric trajectory from temporal progression to resolve the temporal compression observed along standard trajectories. Specifically, we construct a principled timestep schedule that paces diffusion steps proportionally to empirical identity decay (Fig.~\ref{fig:llaco_pipeline}~(c)).

\noindent \textbf{Logarithmic rank retention.}
As established in Proposition~\ref{prop:ordering_retention}, top-1 retention $P$ ignores post-overtaking dynamics, whereas linear retention $Q$ is overly diluted by the vast vocabulary tail. 
We therefore adopt logarithmic rank retention $Q_{\log}$ as our principled schedule.
To capture the true embedding geometry without idealized Gaussian assumptions, we evaluate $Q_{\log}(\gamma)$ by uniformly sampling source tokens across the vocabulary under the spherical noising process (Sec.~\ref{sec:spherical_noising}). 
Following isotonic regression, we fit an endpoint-normalized Logistic function with slope $k$ and center $\gamma_c$:
\begin{equation}
\begin{gathered}
    S(\gamma) = \big(1 + \exp[k(\gamma - \gamma_c)]\big)^{-1}, \\
    Q_{\mathrm{fit}}(\gamma) = \frac{S(\gamma) - S(1)}{S(0) - S(1)}.
\end{gathered}
\label{eq:qlog_logistic}
\end{equation}
This formulation enforces the boundary conditions $Q_{\mathrm{fit}}(0) = 1$ and $Q_{\mathrm{fit}}(1) = 0$ (Appendix~\ref{app:curve_fitting}).

\noindent \textbf{Semantic time mapping.}
We define semantic time $t_{\mathrm{sem}} \in [0, 1]$ by coupling the fitted decay curve with a regularizing linear term:
\begin{equation}
    t_{\mathrm{sem}}(\gamma) = (1 - \epsilon)\big[1 - Q_{\mathrm{fit}}(\gamma)\big] + \epsilon\gamma,
    \label{eq:semantic_time}
\end{equation}
where $\epsilon = 0.1$. This satisfies:
\begin{equation}
    \frac{dt_{\mathrm{sem}}}{d\gamma} = -(1 - \epsilon) Q_{\mathrm{fit}}'(\gamma) + \epsilon \ge \epsilon > 0,
\end{equation}
guaranteeing that $t_{\mathrm{sem}}(\gamma)$ is strictly increasing, smooth, and globally invertible.
Sampling uniformly in semantic time $t_{\mathrm{sem}} \sim \operatorname{Unif}[0, 1]$ induces the sampling density $dt_{\mathrm{sem}} / d\gamma$ along the geometric trajectory $\gamma$.
Consequently, regions of rapid rank decay (where $|Q_{\mathrm{fit}}'(\gamma)|$ peaks) are allocated significantly more training iterations and sampling budget. 
Meanwhile, the linear term $\epsilon\gamma$ provides a density floor of $\epsilon$, so early and nearly flat retention phases retain non-zero coverage.

\subsection{Blockwise Denoising}
\label{sec:blockwise_denoising}
LLaCO generates text through sequential blockwise denoising conditioned on preceding clean context. 
Each target block is initialized at the mask-biased noisy endpoint and iteratively denoised along a uniform grid in semantic time $t_{\mathrm{sem}}$ (Fig.~\ref{fig:llaco_pipeline}~(c)), whose corresponding geometric coordinates $1 = \gamma_0 > \gamma_1 > \dots > \gamma_{N-1} > 0$ are obtained by inverting Eq.~\eqref{eq:semantic_time}. 
At each intermediate step $k$, the network's predicted logits yield a clean-direction estimate $\widehat{x}_{0,k}$ via a probability-weighted combination of leading token directions. 
Because the forward geodesic interpolation (Eq.~\eqref{eq:spherical_path}) preserves constant-speed angular motion, stepping from geometric coordinate $\gamma_k$ toward clean target $\widehat{x}_{0,k}$ to reach $\gamma_{k+1}$ requires traversing an angular ratio of $\frac{\gamma_k - \gamma_{k+1}}{\gamma_k} = 1 - \frac{\gamma_{k+1}}{\gamma_k}$. 
Accordingly, the continuous state is updated along the great-circle arc via SLERP:
\begin{equation}
    s_{k+1} = \operatorname{SLERP}\!\left(s_k,\, \widehat{x}_{0,k};\, 1 - \frac{\gamma_{k+1}}{\gamma_k}\right),
    \label{eq:main_reverse_update}
\end{equation}
which strictly tracks the reverse noising trajectory whenever $\widehat{x}_{0,k}$ coincides with the clean direction at every step. 
Upon completing $N$ denoising iterations, tokens for the current block are decoded by greedy selection ($\operatorname{argmax}$) over the predicted logits and appended as clean context for subsequent blocks (see Appendix~\ref{app:sampling_procedure}).

\begin{table*}[!t]
\centering
\small
\setlength{\tabcolsep}{5.5pt}
\renewcommand{\arraystretch}{0.95}
\caption{Overall performance comparison and ablations. Models are grouped by generation paradigm, with LLaCO-8B listed in both groups for reference. Indented rows under LLaCO-8B alter a single factor of the default configuration and are retrained. Bold values denote the best result within each primary group.}
\label{tab:overall_ablation}
\begin{tabular}{@{}l*{9}{c}@{}}
\toprule
Model / configuration & ARC-E & ARC-C & MMLU & GPQA & BBH & GSM8K & MATH & HumanEval & MBPP \\
\midrule
\multicolumn{10}{@{}l}{\textit{Discrete diffusion}} \\
LLaDA-8B & 75.08 & 53.67 & 63.97 & 26.79 & 53.16 & 78.70 & 43.82 & 45.12 & 40.20 \\
LLaDA-1.5-8B & 79.46 & 55.20 & 64.19 & 27.01 & 52.72 & 79.38 & 44.26 & 46.34 & 43.20 \\
Dream-7B & \textbf{83.50} & \textbf{55.29} & 69.74 & 30.13 & \textbf{54.68} & 80.14 & 51.20 & \textbf{54.27} & \textbf{57.80} \\
\textbf{LLaCO-8B} & 76.68 & 54.78 & \textbf{69.89} & \textbf{31.25} & 51.30 & \textbf{80.74} & \textbf{62.12} & 48.17 & 46.20 \\
\midrule 
\multicolumn{10}{@{}l}{\textit{Continuous diffusion}} \\
TESS-2-7B & 66.58 & 38.05 & 47.98 & 30.13 & 37.05 & 23.96 & 4.70 & 1.22 & 8.60 \\
DSL-LLaDA-8B & 73.15 & 48.81 & 59.05 & 28.35 & 46.81 & 25.40 & 15.10 & 20.73 & 25.60 \\
\textbf{LLaCO-8B} & \textbf{76.68} & \textbf{54.78} & \textbf{69.89} & \textbf{31.25} & \textbf{51.30} & \textbf{80.74} & \textbf{62.12} & \textbf{48.17} & \textbf{46.20} \\
\cmidrule(l{1em}){1-10}
\quad $P$ schedule & 70.12 & 52.30 & 61.77 & 29.69 & 46.72 & 64.82 & 35.00 & 35.98 & 35.60 \\
\quad $Q$ schedule & 72.43 & 53.84 & 65.74 & 30.80 & 48.61 & 77.56 & 58.30 & 40.24 & 37.20 \\
\quad w/o mask-biased endpoint & 74.83 & 54.01 & 67.28 & 30.58 & 48.16 & 79.23 & 60.02 & 30.49 & 36.60 \\
\quad w/o spherical noising & 74.58 & 54.58 & 67.35 & 31.03 & 49.09 & 78.39 & 59.22 & 43.29 & 41.40 \\
\bottomrule
\end{tabular}
\vspace{-2mm}
\end{table*}

\section{Experiments}
\label{sec:experiments}

\subsection{Experimental Setup}
\label{sec:experimental_setup}
We instantiate LLaCO by adapting an autoregressive--diffusion hybrid backbone~\cite{fu2026nemotron}. 
The original token embeddings remain frozen, while the embedding directional adapter, timestep-conditioned modulation, and transformer backbone are optimized end-to-end using Adam~\cite{kingma2015adam}. 
Generation employs blockwise denoising with block size $B = 4$ and $N = 4$ steps along the reallocated schedule (Appendices~\ref{app:sampling_procedure} and~\ref{app:block_steps}). 
The model is trained on 1.32M open-source multi-domain instruction examples~\cite{nathawani2025nemotronpost,nvidia2025nemotronnano2} (detailed data configurations in Appendix~\ref{app:training_sampling}).

\subsection{Quantitative Evaluation Metrics}
We benchmark model capabilities across three core domains. 
\textit{(i) Knowledge and reasoning} covers ARC-Easy, ARC-Challenge~\cite{clark2018think}, MMLU~\cite{hendrycks2021mmlu}, GPQA~\cite{rein2024gpqa}, and BBH~\cite{suzgun2023bbh}. 
\textit{(ii) Mathematical problem solving} is evaluated on GSM8K~\cite{cobbe2021gsm8k} and MATH~\cite{hendrycks2021math}. 
\textit{(iii) Code generation} includes HumanEval~\cite{chen2021humaneval} and MBPP~\cite{austin2021program}. 
All evaluations are executed within the NeMo-Skills framework~\cite{nvidia2024nemoskills}.

\subsection{Comparison with Diffusion-based LLMs}
\label{sec:overall_comparison}

We compare LLaCO with continuous diffusion language models (TESS-2~\cite{tae2025tess2}, DSL-LLaDA~\cite{yu2026dslllada}) and discrete diffusion baselines (LLaDA~\cite{nie2025large}, LLaDA-1.5~\cite{zhu2025llada15}, Dream~\cite{ye2025dream}).
All evaluated models operate at comparable 7B to 8B parameter scales to ensure fair comparison (Appendix~\ref{app:baseline_selection}). 
As shown in Table~\ref{tab:overall_ablation}, LLaCO comprehensively outperforms all continuous baselines across every evaluated benchmark, demonstrating decisive advantages in mathematical problem solving and code generation. Specifically, it yields absolute improvements of 47.02 points on MATH and 27.44 points on HumanEval over the strongest continuous baseline. 
Against representative discrete diffusion models, LLaCO exhibits competitive performance across knowledge, reasoning, and code generation. While the discrete baseline Dream-7B retains a 6.10-point advantage on HumanEval, LLaCO presents marginal leads on MMLU and GPQA, and a significant 10.92-point lead on MATH.

\subsection{Ablation Study}
\label{sec:ablations_sampling}

To validate the effectiveness of our principled schedule and key design components, we construct four ablation variants, reported in Table~\ref{tab:overall_ablation}.
Here, the $P$ and $Q$ pacing schedules are derived from their respective empirical retention fits, while the unbiased endpoint variant remeasures and refits $Q_{\log}$ on isotropic noise (Appendix~\ref{app:curve_fitting}).


\noindent \textbf{P schedule.}
We substitute the schedule $Q_{\log}$ with the top-1 retention schedule $P$. Focusing solely on first-rank transitions compresses the subsequent rank decay into a narrow terminal window, causing the model to misallocate its optimization budget. Consequently, this pacing choice degrades performance in mathematical reasoning and code generation.

\noindent \textbf{Q schedule.}
We replace $Q_{\log}$ with the linear rank retention schedule $Q$. 
While $Q$ captures global rank decay across the vocabulary, it is diluted by the vast token tail and under-allocates optimization budget to competitive leading ranks. 
Consequently, this schedule achieves suboptimal overall performance compared to $Q_{\log}$.

\noindent \textbf{W/o Mask-biased endpoint.}
We set $\lambda = 0$ to adopt isotropic terminal noise. 
Without this semantic anchor, the diffusion trajectory fails to exploit the inductive bias of the pretrained discrete LM, impeding knowledge transfer from the base model. 
Thus, this variant leads to performance drops in code generation.

\noindent \textbf{W/o Spherical noising.}
We replace the spherical geodesic path with standard Euclidean linear noising. 
Under Euclidean corruption, embedding norm decay causes spurious overtaking and prevents tokens with varying norms from sharing an accurate competition scale. 
Consequently, this misaligned competition distorts the trajectory pacing and degrades performance across all evaluated tasks.

\begin{table}[t]
\centering
\small
\setlength{\tabcolsep}{0pt}
\newcommand{\rep}[1]{\textcolor{gray}{#1}}
\caption{Effect of inference block size $B$ and steps per block $N$ with the default LLaCO-8B checkpoint. Bold marks the default setting, which is best on all tasks; the gray row repeats $(8,8)$ for reference.}
\label{tab:block_steps}
\begin{tabular*}{\textwidth}{@{\extracolsep{\fill}}ccccccccccc@{}}
\toprule
\multicolumn{5}{c}{\textit{Varying block size} ($N=B$)} & \hspace{1.2em} & \multicolumn{5}{c}{\textit{Varying steps} ($B=8$)}\\
\cmidrule{1-5}\cmidrule{7-11}
$B$ & GSM8K & MATH & HumanEval & MBPP && $N$ & GSM8K & MATH & HumanEval & MBPP\\
\midrule
\textbf{4} & \textbf{80.74} & \textbf{62.12} & \textbf{48.17} & \textbf{46.20} && 4 & 67.32 & 45.80 & 28.66 & 32.60\\
8 & 70.96 & 47.70 & 31.71 & 33.40 && \rep{8} & \rep{70.96} & \rep{47.70} & \rep{31.71} & \rep{33.40}\\
16 & 48.07 & 24.28 & 15.24 & 20.20 && 16 & 67.48 & 44.76 & 34.15 & 31.40\\
\bottomrule
\end{tabular*}
\end{table}

\subsection{Sampling Analysis}


Beyond these retrained variants, we examine how inference-time sampling parameters influence the quality of LLaCO.
Table~\ref{tab:block_steps} evaluates the impact of block size $B$ and denoising steps per block $N$.

\noindent \textbf{Block size.}
We vary the block size $B \in \{4, 8, 16\}$ under a synchronized step setting $N = B$. 
Because tokens within a block are committed simultaneously at the terminal denoising step, their mutual contextual dependencies must be resolved jointly through the continuous state trajectory. 
Larger blocks force more interdependent tokens into this simultaneous continuous resolution, substantially compounding reasoning errors. 
Consequently, expanding the block size monotonically degrades generation quality across all evaluated tasks.

\noindent \textbf{Steps per block.}
Fixing $B = 8$, we vary the denoising steps per block $N \in \{4, 8, 16\}$. 
Additional steps provide finer discretization to refine the clean-direction estimate $\widehat{x}_{0,k}$. 
However, each intermediate estimation error can divert the state away from the nominal geodesic path in Eq.~\eqref{eq:main_reverse_update}. 
As a result, increasing $N$ from 4 to 8 yields consistent improvements across all benchmarks, whereas further increasing $N$ to 16 causes task-dependent fluctuations, demonstrating a trade-off
between semantic refinement and trajectory drift.

\section{Conclusion}
\label{sec:conclusion}
In this work, we investigate the fundamental token competition dynamics underlying continuous diffusion language models. 
Through our Competition Threshold Theory, we reveal how Euclidean norm decay induces spurious overtaking, while high-dimensional geometry causes temporal compression of identity decay under standard diffusion schedules, motivating our Semantic Retention Scheduling based on logarithmic rank retention.
Building upon these insights, we introduce LLaCO, a principled framework featuring spherical geodesic trajectories, an embedding directional adapter, timestep-conditioned modulation, and mask-biased endpoints. 
Extensive evaluations demonstrate that LLaCO outperforms all evaluated continuous diffusion baselines and achieves competitive performance against discrete diffusion models of comparable scale. 
Together, these theoretical insights and architectural designs provide a principled foundation for future advancements in continuous diffusion language modeling.

\section*{Limitations}
Some of our theoretical conclusions rely on near-orthogonal structures and related statistical approximations in high-dimensional and sparsely populated representation spaces. Their applicability to alternative representation geometries and noise distributions warrants further validation. 

\clearpage
\bibliographystyle{IEEEtran}
\bibliography{custom}

\clearpage
\beginappendix
\numberwithin{equation}{section}


\section{Details of Theoretical Analysis}
\label{app:theory}

This appendix follows the order of Section~\ref{sec:theoretical_analysis}. Appendices~\ref{app:euclidean_pathology}--\ref{app:gaussian_competition} support Section~\ref{sec:CTT}: directional identity, competition thresholds, and pairwise overtaking. Appendices~\ref{app:rank_normalization}, \ref{app:first_overtaking}, and \ref{app:path_factorization} support Section~\ref{sec:SRS}: retention statistics, the first-overtaking scale, and trajectory factorization. Appendices~\ref{app:source_projection}--\ref{app:near_orthogonal} support Section~\ref{sec:val}: spectrum-aware and near-orthogonal predictions. Appendix~\ref{app:preservation_shared} collects preservation bounds, which support the preservation guarantee in Section~\ref{sec:CTT}, and shared-component scaling, which complements the main text but is not used in its derivations.

Throughout, $P_{\mathrm{raw}}(\kappa)=\Pr[r(\kappa)=1]$ denotes raw (unnormalized) top-1 retention and $P$ its endpoint-normalized version (Appendix~\ref{app:rank_normalization}). Trajectory position $\gamma\in[0,1]$ increases from the clean state ($\gamma=0$) to the noisy endpoint ($\gamma=1$).

\subsection{Spurious Overtaking under Embedding Norm Decay}
\label{app:euclidean_pathology}

This section supports the directional-identity argument in Section~\ref{sec:directional_identity}: under Euclidean ranking, embedding norm decay alone can cause spurious overtaking.

Consider the noiseless trajectory $s(\alpha)=\alpha e_i$ with $\alpha\in(0,1]$, i.e., $\beta=0$. For any competitor $j\ne i$, let $n_j=\|e_j\|>0$, $u_j=e_j/n_j$, and $a_{ij}=u_i^\top u_j<1$. The difference between the squared Euclidean distances to candidate $j$ and to the source is
\begin{equation}
\begin{aligned}
    G_{ij}(\alpha)
    &= \|\alpha e_i - e_j\|^2 - \|\alpha e_i - e_i\|^2 \\
    &= \big(\alpha^2 n_i^2 - 2\alpha e_i^\top e_j + n_j^2\big) \\
    &\quad{}- (1 - \alpha)^2 n_i^2 \\
    &= 2\alpha n_i (n_i - n_j a_{ij}) + n_j^2 - n_i^2.
\end{aligned}
\label{eq:app_euclidean_gap}
\end{equation}
Candidate $j$ overtakes the source in Euclidean distance if and only if $G_{ij}(\alpha)<0$. The function $G_{ij}$ is affine in $\alpha$, with $G_{ij}(1)=\|e_i-e_j\|^2>0$ and $G_{ij}(0)=n_j^2-n_i^2$.
If $n_j\ge n_i$, then $G_{ij}(0)\ge0$ and $G_{ij}(1)>0$, so $G_{ij}(\alpha)>0$ for all $\alpha\in(0,1]$. If $n_j<n_i$, then $G_{ij}(0)<0<G_{ij}(1)$, so $G_{ij}$ is increasing and has a unique zero
\begin{equation}
    \alpha^0_{ij} = \frac{n_i^2 - n_j^2}{2n_i(n_i - n_j a_{ij})}\in(0,1).
    \label{eq:app_alpha_zero}
\end{equation}
Hence candidate $j$ overtakes the noiseless state if and only if $n_j<n_i$ and $\alpha<\alpha^0_{ij}$. In contrast, $s(\alpha)$ has direction $u_i$ for every $\alpha>0$, so its cosine score is $c_i=1>a_{ij}=c_j$ and cosine ranking keeps the source at rank 1. Euclidean overtaking can therefore arise from norm differences alone, without any directional corruption.

\subsection{Competition Thresholds}
\label{app:threshold_details}

This section proves Proposition~\ref{prop:competition_thresholds} and fixes the boundary conventions used in the rest of the appendix.

\paragraph{Conventions.}
We use the notation of Section~\ref{sec:token_competition}. All embeddings and evaluated states are nonzero. Rankings count candidates whose cosine similarity is strictly greater than that of the source, so a tie does not count as overtaking. Identical directions are combined into one direction class, so distinct candidates satisfy $a_{ij}<1$. The candidate count $V$ is the size of the resulting ranking vocabulary, and the source distribution $\pi$ is understood on these classes if any are merged.

\paragraph{Identity decay as directional competition.}
\label{app:information_competition}
Let the source be uniform over the $V$ candidates with unit directions ($n_j=1$, so $\kappa=\beta/\alpha$), and let $s=\alpha u_i+\beta z$ with $\beta>0$ and $z\sim\mathcal N(0,I_d/d)$. Conditioned on source $j$, $s\sim\mathcal N(\alpha u_j,(\beta^2/d)I_d)$. These class-conditional densities share an isotropic covariance and have equal-norm means, so the posterior is a softmax of linear scores \citep[\S4.2.1]{bishop2006pattern}:
\begin{equation}
\begin{aligned}
\ell_{ij}=\ln\frac{p(j\mid s)}{p(i\mid s)}
&=\frac{d\alpha}{\beta^2}(u_j-u_i)^\top s\\
&=\frac{d}{\kappa^2}\big[-(1-a_{ij})+\kappa(b_j-b_i)\big].
\end{aligned}
\label{eq:app_posterior_odds}
\end{equation}
This is the bracket of Eq.~\eqref{eq:competition_gap} up to a positive factor. Hence candidate $j$ overtakes the source exactly when the posterior prefers $j$, $r_i(\kappa)$ is the posterior rank of the source, and the source's posterior mass $p(i\mid s)=\big[1+\sum_{j\ne i}e^{\ell_{ij}}\big]^{-1}$ is determined by the pairwise directional competition terms alone. By the coupling in Appendix~\ref{app:gaussian_competition}, these rank statements carry over approximately to spherical endpoints.

\paragraph{Proof of Proposition~\ref{prop:competition_thresholds}.}
For $\alpha>0$, the positive factor $\alpha n_i/\|s\|$ in Eq.~\eqref{eq:competition_gap} does not affect the sign of the score difference. If $b_j-b_i\le0$, then
\begin{equation}
 -(1-a_{ij})+\kappa(b_j-b_i)<0
\end{equation}
for every finite $\kappa\ge0$, so candidate $j$ never overtakes. Otherwise, dividing by $b_j-b_i>0$ gives the threshold in Eq.~\eqref{eq:pairwise_threshold}. Summing the strict overtaking indicators proves Eq.~\eqref{eq:rank_after_overtaking}; each indicator is non-decreasing in $\kappa$, so the rank is non-decreasing in $\kappa$ for each fixed $(i,z)$. Define
\begin{equation}
 M_{i,z}=\max_{j\ne i}\frac{(b_j-b_i)_+}{1-a_{ij}}.
\end{equation}
Then $\kappa^*_{i,z}=1/M_{i,z}$ when $M_{i,z}>0$, and $\kappa^*_{i,z}=\infty$ otherwise. Under the strict-overtaking convention, the source remains top-1 at equality $\kappa=\kappa^*_{i,z}$. This yields $P_{\mathrm{raw}}(\kappa)=\Pr[\kappa^*\ge\kappa]$.

\paragraph{Noisy endpoint.}
When the noisy endpoint has $\alpha=0$ and $\beta>0$, its ranking is defined directly by $b_j$. Since the vocabulary is finite, the limiting rank as $\kappa\to\infty$ counts exactly those candidates with $b_j>b_i$, matching this endpoint convention. Samples with no such candidate have terminal rank 1.

\subsection{Pairwise Overtaking under Gaussian Noise}
\label{app:gaussian_competition}

This section derives Eq.~\eqref{eq:pairwise_prob} and the corresponding expected rank.

\paragraph{Expected rank.}
For any noise distribution, taking the expectation of Eq.~\eqref{eq:rank_after_overtaking} and using linearity gives Eq.~\eqref{eq:expected_rank}.

\paragraph{Pairwise overtaking probability.}
Let $z\sim\mathcal N(0,I_d/d)$, so that $\mathbb E\|z\|^2=1$ and the noise has the same scale as a unit token direction. Any projection of $z$ onto a fixed vector is Gaussian. Since $\|u_j-u_i\|^2=2(1-a_{ij})$, the relative noise advantage $D_{ij}=z^\top(u_j-u_i)=b_j-b_i$ has mean zero and variance
\begin{equation}
    \operatorname{Var}(D_{ij}) = \frac{\|u_j - u_i\|^2}{d} = \frac{2(1 - a_{ij})}{d}.
\end{equation}
For $\kappa>0$, the overtaking condition in Eq.~\eqref{eq:pairwise_threshold} is $D_{ij}>(1-a_{ij})/\kappa$. Standardizing gives Eq.~\eqref{eq:pairwise_prob}:
\begin{equation}
 p_{ij}(\kappa)=\bar\Phi\!\left(\frac{1}{\kappa}\sqrt{\frac{d(1-a_{ij})}{2}}\right).
\label{eq:app_pairwise_probability}
\end{equation}
At $\kappa=0$, $p_{ij}(0)=0$. The probability increases with both $\kappa$ and $a_{ij}$. For a fixed source,
\begin{equation}
    \mathbb{E}[r_i(\kappa)] = 1 + \sum_{j \ne i} p_{ij}(\kappa),
    \label{eq:app_expected_rank}
\end{equation}
and averaging over sources gives the Gaussian form of Eq.~\eqref{eq:expected_rank}.

\paragraph{Spherical noise.}
Uniform spherical noise can be coupled to the Gaussian model. Using the same sample $z_G\sim\mathcal N(0,I_d/d)$, set $R_G=\|z_G\|$ and $z_S=z_G/R_G$, which is uniform on the unit sphere and independent of $R_G$. Every relative noise advantage is divided by $R_G$, so samplewise
\begin{equation}
 \tau_{ij}^{S}=R_G\,\tau_{ij}^{G},\qquad
 \kappa_{i,S}^*=R_G\,\kappa_{i,G}^*,
 \label{eq:app_spherical_coupling}
\end{equation}
and $r_{i,z_S}(\kappa)=r_{i,z_G}(\kappa/R_G)$. For $0<\epsilon<1$, let $\delta_\epsilon=\Pr(|R_G-1|>\epsilon)$. On the complementary event, $\kappa_{i,S}^*$ lies between $(1-\epsilon)\kappa_{i,G}^*$ and $(1+\epsilon)\kappa_{i,G}^*$, which yields
\begin{equation}
 \begin{aligned}
 P_S(\kappa)&\ge
 P_G\!\left(\frac\kappa{1-\epsilon}\right)-\delta_\epsilon,\\
 P_S(\kappa)&\le
 P_G\!\left(\frac\kappa{1+\epsilon}\right)+\delta_\epsilon,
 \end{aligned}
 \label{eq:app_spherical_sandwich}
\end{equation}
where $P_G$ and $P_S$ denote raw top-1 retention under the two noise models with the same source distribution. The terminal ranks, and hence the valid set $E$, depend only on $z_S$ and coincide for both models. Since per-trajectory scores are nonincreasing in $\kappa$ and $R_G$ is independent of $z_S$, the same sandwich holds for $P$, $Q$, and $Q_{\log}$. Because $R_G$ concentrates around 1, $\delta_\epsilon\to0$ as $d$ grows for any fixed $\epsilon$, so Gaussian-model predictions carry over to spherical endpoints. Fig.~\ref{fig:prediction_measurement}(a) compares Eq.~\eqref{eq:app_expected_rank} with the expected ranks measured under both Gaussian and spherical endpoints (Appendix~\ref{app:embedding_validation}).

\subsection{Retention Statistics and Their Ordering}
\label{app:rank_normalization}

This section supports the retention statistics in Section~\ref{sec:rank_statistics} and proves Proposition~\ref{prop:ordering_retention}. The results require only a non-decreasing rank along the trajectory, clean rank 1, and $\Pr(E)>0$; they hold for any source and noise distribution.

\paragraph{Endpoint normalization.}
Along a trajectory with non-decreasing $\kappa(\gamma)$, Proposition~\ref{prop:competition_thresholds} gives a non-decreasing count $M(\gamma)=r(\gamma)-1$ with $M(0)=0$. Let $M_{\mathrm{end}}=M(1)$ and $E=\{M_{\mathrm{end}}>0\}$. On $E^c$, monotonicity forces $M(\gamma)=0$ for all $\gamma$. Hence $P_{\mathrm{raw}}(\gamma)=\Pr(E^c)+\Pr(E,\,M(\gamma)=0)$ and $P_{\mathrm{raw}}(1)=\Pr(E^c)$, so
\begin{equation}
 \begin{aligned}
 P(\gamma)&=\mathbb E\big[\mathbb I\{M(\gamma)=0\}\mid E\big]\\
 &=\frac{P_{\mathrm{raw}}(\gamma)-P_{\mathrm{raw}}(1)}{1-P_{\mathrm{raw}}(1)}.
 \end{aligned}
 \label{eq:app_p_normalization}
\end{equation}
Since $P(\gamma)-P_{\mathrm{raw}}(\gamma)=-P_{\mathrm{raw}}(1)[1-P_{\mathrm{raw}}(\gamma)]/[1-P_{\mathrm{raw}}(1)]$ and $P_{\mathrm{raw}}(\gamma)\ge P_{\mathrm{raw}}(1)$, $P$ and $P_{\mathrm{raw}}$ differ by at most $P_{\mathrm{raw}}(1)$. This terminal value equals $1/V$ in the near-orthogonal surrogate (Appendix~\ref{app:near_orthogonal}) and is about $10^{-5}$ in our measurements (Appendix~\ref{app:embedding_validation}), so the main text does not distinguish $P$ from $P_{\mathrm{raw}}$ when stating approximations.

The per-trajectory scores in Eq.~\eqref{eq:trajectory_retention} are endpoint normalizations of rank scores. With
\begin{equation}
 \begin{aligned}
 K(\gamma)&=1-\frac{r(\gamma)-1}{V-1},\\
 K_{\log}(\gamma)&=1-\frac{\ln(1+r(\gamma))}{\ln(1+V)},
 \end{aligned}
\end{equation}
direct substitution on $E$ gives
\begin{equation}
 \begin{aligned}
 \frac{K(\gamma)-K(1)}{K(0)-K(1)}&=q(\gamma),\\
 \frac{K_{\log}(\gamma)-K_{\log}(1)}{K_{\log}(0)-K_{\log}(1)}&=q_{\log}(\gamma).
 \end{aligned}
\end{equation}
Endpoint normalization thus removes the dependence on $V$, and $Q$, $Q_{\log}$ average these scores over $E$.

\paragraph{Logarithmic decomposition.}
For integer $1\le r\le V$, telescoping gives
\begin{equation}
 \begin{aligned}
 \sum_{q=2}^{V}\ln\frac{q+1}{q}\,\mathbb I\{r\ge q\}
 &=\sum_{q=2}^{r}\ln\frac{q+1}{q}\\
 &=\ln\frac{1+r}{2}.
 \end{aligned}
\end{equation}
The sum is empty at $r=1$. The weights satisfy $1/(q+1)<\ln((q+1)/q)<1/q$, so crossing rank level $q$ costs approximately $1/q$, and $q_{\log}$ emphasizes the leading ranks.

\paragraph{Proof of the ordering.}
Let $g(n)=\ln(1+n/2)$, so that $\ln\frac{1+r}{2}=g(r-1)$ and, on $E$,
\begin{equation}
 \begin{aligned}
 q(\gamma)&=1-\frac{M(\gamma)}{M_{\mathrm{end}}},\\
 q_{\log}(\gamma)&=1-\frac{g(M(\gamma))}{g(M_{\mathrm{end}})}.
 \end{aligned}
\end{equation}
For $0\le M\le M_{\mathrm{end}}$, write $M=(1-\lambda)\cdot0+\lambda M_{\mathrm{end}}$ with $\lambda=M/M_{\mathrm{end}}$. Concavity of $g$ and $g(0)=0$ give $g(M)\ge\lambda g(M_{\mathrm{end}})$, hence $q_{\log}(\gamma)\le q(\gamma)$. Since $g$ is increasing, $q_{\log}(\gamma)\ge0$, with equality to 1 when $M(\gamma)=0$. Therefore
\begin{equation}
 \mathbb I\{M(\gamma)=0\}\le q_{\log}(\gamma)\le q(\gamma).
\end{equation}
Averaging over $E$ yields Eq.~\eqref{eq:rank_statistic_order} for every $\gamma\in[0,1]$.

\paragraph{Count thresholds.}
For $\eta\in(0,1)$, $q_{\log}(\gamma)\le\eta$ means $g(M(\gamma))\ge(1-\eta)\,g(M_{\mathrm{end}})$. Since $g$ is strictly increasing and $g(M_{\mathrm{end}})>0$ on $E$, this is equivalent to
\begin{equation}
 M(\gamma)\ge2\big[(1+M_{\mathrm{end}}/2)^{1-\eta}-1\big],
\end{equation}
while $q(\gamma)\le\eta$ if and only if $M(\gamma)\ge(1-\eta)M_{\mathrm{end}}$; exact integer thresholds round these expressions upward. This completes the proof of Proposition~\ref{prop:ordering_retention}. At $\eta=1/2$, the logarithmic threshold is $2\sqrt{1+M_{\mathrm{end}}/2}-2=\sqrt{2M_{\mathrm{end}}}-2+\mathcal O(M_{\mathrm{end}}^{-1/2})$, versus $M_{\mathrm{end}}/2$ for the linear score. More generally, for fixed $\eta$ and large $M_{\mathrm{end}}$, the logarithmic threshold grows as $2^{\eta}M_{\mathrm{end}}^{1-\eta}$, whereas the linear threshold grows as $M_{\mathrm{end}}$.

Thus, when $M_{\mathrm{end}}$ is large, $q_{\log}$ reaches a given level after only a vanishing fraction, of order $M_{\mathrm{end}}^{-\eta}$, of the terminal candidates has overtaken; unlike the top-1 indicator, it continues to decrease until all $M_{\mathrm{end}}$ crossings occur.

\subsection{Retention Spectra under Gaussian Noise}
\label{app:source_projection}
\label{app:qlog_conditional}

This section derives the source-projection decomposition, the exact conditional count in Eq.~\eqref{eq:conditional_count}, and the spectrum predictions in Eq.~\eqref{eq:spectrum_predictions}, and bounds the error of replacing counts by their conditional means. Let $z\sim\mathcal N(0,I_d/d)$ and $\sigma_b=d^{-1/2}$.

\paragraph{Coupling through the source projection.}
The expected rank depends only on marginal overtaking probabilities, but $P$, $Q$, and $Q_{\log}$ depend on the joint behavior of candidates. For distinct candidates $j\ne k$ ($j,k\ne i$), the relative noise advantages share the term $b_i$:
\begin{equation}
\begin{aligned}
    &\operatorname{Cov}(D_{ij},D_{ik}) \\
    &\quad=\frac{(u_j-u_i)^\top(u_k-u_i)}{d} \\
    &\quad=\frac{1+a_{jk}-a_{ij}-a_{ik}}{d},
\end{aligned}
\end{equation}
so that
\begin{equation}
    \begin{aligned}
    &\operatorname{Corr}(D_{ij},D_{ik})\\
    &\quad=\frac{1+a_{jk}-a_{ij}-a_{ik}}
    {2\sqrt{(1-a_{ij})(1-a_{ik})}}.
    \end{aligned}
    \label{eq:app_noise_correlation}
\end{equation}
For nearly orthogonal directions ($a_{ij},a_{ik},a_{jk}\to0$), this correlation approaches $1/2$: a noise sample that lowers $b_i$ raises the relative advantage of every candidate simultaneously. Conditioning on the source projection removes this common component.

\paragraph{Source-projection decomposition.}
For a candidate with $-1<a_{ij}<1$, decompose $u_j=a_{ij}u_i+\sqrt{1-a_{ij}^2}\,w_j$, where $w_j^\top u_i=0$ and $\|w_j\|=1$. Define $Y=z^\top u_i/\sigma_b$ and $Z_j=z^\top w_j/\sigma_b$. Since each $w_j$ is orthogonal to $u_i$, $Y\sim\mathcal N(0,1)$ is independent of $(Z_j)_{j\ne i}$; each $Z_j$ is standard Gaussian, with covariance
\begin{equation}
 C_{jk}=\frac{a_{jk}-a_{ij}a_{ik}}
 {\sqrt{(1-a_{ij}^2)(1-a_{ik}^2)}}.
 \label{eq:app_residual_cov}
\end{equation}
The relative noise advantage is
\begin{equation}
 b_j-b_i=\sigma_b\big[\sqrt{1-a_{ij}^2}\,Z_j-(1-a_{ij})Y\big].
\end{equation}
With $h=1/(\kappa\sigma_b)$ and $\rho_{ij}=\sqrt{(1-a_{ij})/(1+a_{ij})}$, dividing the overtaking condition $\kappa(b_j-b_i)>1-a_{ij}$ by $\kappa\sigma_b\sqrt{1-a_{ij}^2}$ gives
\begin{equation}
 Z_j>\rho_{ij}(Y+h).
 \label{eq:app_overtaking_condition}
\end{equation}
An antipodal candidate ($a_{ij}=-1$) overtakes if and only if $Y<-h$, the $\rho_{ij}\to\infty$ limit of Eq.~\eqref{eq:app_overtaking_condition}. Below we assume $a_{ij}>-1$; an antipodal candidate would add the event $\{Y<-h\}$ to the overtaking events.

\paragraph{Top-1 retention.}
The source remains top-1 when no candidate satisfies Eq.~\eqref{eq:app_overtaking_condition}:
\begin{equation}
 \begin{gathered}
 P_{\mathrm{raw}}(\kappa\mid i)
 =\int_{-\infty}^{\infty}\!\phi(y)\\
 {}\times\Phi_C\!\left(\{\rho_{ij}(y+h)\}_{j\ne i}\right)dy,
 \end{gathered}
 \label{eq:app_correlated_preservation}
\end{equation}
where $\Phi_C$ is the CDF of the zero-mean Gaussian vector with covariance $C$. Because $\Phi_C$ has dimension comparable to $V$, we approximate the residuals as independent standard Gaussians:
\begin{equation}
 \begin{aligned}
 &P_{\mathrm{raw}}(\kappa\mid i)\\
 &\quad\approx\int_{-\infty}^{\infty}\!\phi(y)
 \prod_{j\ne i}\Phi\!\left(\rho_{ij}(y+h)\right)dy.
 \end{aligned}
 \label{eq:app_spectrum_approx}
\end{equation}
This is the first line of Eq.~\eqref{eq:spectrum_predictions}. It retains each actual similarity $a_{ij}$. Under this approximation, overtaking events are independent given $Y$ but remain dependent after averaging over the shared $Y$. Its value at $h=0$ gives $P_{\mathrm{raw}}(\infty\mid i)$, from which $P$ follows by Eq.~\eqref{eq:app_p_normalization}.

\paragraph{Conditional overtaking counts.}
On a trajectory with non-decreasing $\kappa(\gamma)$ that reaches $\kappa=\infty$ at the endpoint, candidate $j$ has overtaken at $\kappa$ iff $Z_j>\rho_{ij}(Y+h)$, and at the endpoint iff $Z_j>\rho_{ij}Y$; since $h\ge0$, current overtaking candidates form a subset of the terminal ones and $M(\gamma)\le M_{\mathrm{end}}$. Because each $Z_j$ is standard Gaussian and independent of $Y$, Eq.~\eqref{eq:conditional_count} holds exactly for any residual covariance $C$. We write $\mu=\mu_i(y,\kappa)$ and $\nu=\mu_i(y,\infty)$, the latter denoted $\mu^T$ in the main text. Using $\int\phi(y)\bar\Phi(\rho(y+h))\,dy=\bar\Phi\big(\rho h/\sqrt{1+\rho^2}\big)$ and $\rho_{ij}/\sqrt{1+\rho_{ij}^2}=\sqrt{(1-a_{ij})/2}$, averaging $\mu$ over $Y$ recovers Eq.~\eqref{eq:app_expected_rank}.

\paragraph{Plug-in predictions.}
By Appendix~\ref{app:rank_normalization},
\begin{equation}
 \begin{aligned}
 Q&=\mathbb{E}\!\left[1-\frac{M(\gamma)}{M_{\mathrm{end}}}\,\middle|\,E\right],\\
 Q_{\log}&=\mathbb{E}\!\left[1-\frac{g(M(\gamma))}{g(M_{\mathrm{end}})}\,\middle|\,E\right].
 \end{aligned}
\end{equation}
Under the independent-residual approximation, define the current and terminal overtaking probabilities and the validity probability
\begin{equation}
 \begin{aligned}
 p_j(y,h)&=\bar\Phi\big(\rho_{ij}(y+h)\big),\\
 p_j^T(y)&=\bar\Phi(\rho_{ij}y),\\
 \omega_i(y)&=1-\prod_{j\ne i}(1-p_j^T(y)).
 \end{aligned}
\end{equation}
Given $(i,Y=y)$, the counts are nested sums of independent Bernoulli pairs with means $\mu=\sum_jp_j$ and $\nu=\sum_jp_j^T$, and $\omega_i(y)=\Pr(E\mid i,Y=y)$. If sources follow $\pi$, the expectation of a function $f(i,Y)$ over valid trajectories is
\begin{equation}
 \mathbb E_E[f]=
 \frac{\sum_i\pi_i\int\phi(y)\omega_i(y)f(i,y)\,dy}
 {\sum_i\pi_i\int\phi(y)\omega_i(y)\,dy}.
 \label{eq:valid_source_projection_weight}
\end{equation}
The $Q$ and $Q_{\log}$ predictions in Eq.~\eqref{eq:spectrum_predictions} replace $M(\gamma)$ and $M_{\mathrm{end}}$ by $\mu$ and $\nu$ inside the ratios and average with this weight. Ratios are evaluated only where $\nu>0$ and $\omega_i(y)>0$; points with $\omega_i(y)=0$ carry zero weight. Since $\mu$ and $\nu$ are conditional on $(i,Y)$ before selecting $E$, the following bounds include the effect of that selection.

\paragraph{Error of the linear ratio.}
Fix $(i,y)$ and the current noise level, abbreviate $M=M(\gamma)$, and write $\omega=\omega_i(y)$. All expectations in this paragraph hold $(i,y)$ fixed. Independence across candidates gives
\begin{equation}
 \begin{aligned}
 \operatorname{Var}(M_{\mathrm{end}})&=\sum_jp_j^T(1-p_j^T)\le\nu,\\
 \operatorname{Cov}(M,M_{\mathrm{end}})&=\sum_jp_j(1-p_j^T)\le\nu.
 \end{aligned}
\end{equation}
On $M_{\mathrm{end}}>0$, expanding the reciprocal gives
\begin{equation}
 \begin{aligned}
 \frac{M}{M_{\mathrm{end}}}&=\frac M\nu-\frac{M(M_{\mathrm{end}}-\nu)}{\nu^2}\\
 &\quad+\frac{M(M_{\mathrm{end}}-\nu)^2}{\nu^2 M_{\mathrm{end}}}.
 \end{aligned}
\end{equation}
Since $M=0$ when $M_{\mathrm{end}}=0$ and $0\le M\le M_{\mathrm{end}}$, averaging yields
\begin{equation}
 \begin{aligned}
 &\mathbb E\!\left[\frac{M}{M_{\mathrm{end}}};M_{\mathrm{end}}>0\right]\\
 &\quad=\frac\mu\nu-\frac{\operatorname{Cov}(M,M_{\mathrm{end}})}{\nu^2}+R,\\
 &0\le R\le\frac1\nu.
 \end{aligned}
\end{equation}
Dividing by $\omega$ and using $0\le\operatorname{Cov}(M,M_{\mathrm{end}})/\nu^2\le1/\nu$,
\begin{equation}
 \begin{aligned}
 &\left|\mathbb E\!\left[\frac{M}{M_{\mathrm{end}}}\middle|M_{\mathrm{end}}>0\right]-\frac\mu\nu\right|\\
 &\quad\le\min\left\{1,\frac{\nu^{-1}+1-\omega}{\omega}\right\}.
 \end{aligned}
 \label{eq:q_ratio_error_bound}
\end{equation}
Because $1-\omega=\prod_j(1-p_j^T)\le e^{-\nu}$, this bound is $\mathcal O(\nu^{-1})$ as $\nu\to\infty$. It is pointwise in $(i,y)$; the curve error is bounded by its average under Eq.~\eqref{eq:valid_source_projection_weight}.

\paragraph{Error of the logarithmic ratio.}
With the same fixed $(i,y)$, write $T=M_{\mathrm{end}}$. We first control the Jensen gap $J(\mu)=g(\mu)-\mathbb E[g(M)]\ge0$. Since $g''(x)=-(2+x)^{-2}$,
\begin{equation}
 0\le J(\mu)\le\operatorname{Var}(M)/8\le\mu/8.
\end{equation}
A bound uniform in $\mu$ also holds. For $0\le\mu\le1$, $J(\mu)\le g(\mu)<1$. For $\mu\ge1$, write $M=\sum_jB_j$ with independent $B_j\sim\operatorname{Bernoulli}(p_j)$; for $u\in[0,1]$, the probability generating function satisfies $\mathbb E[u^M]=\prod_j(1-p_j(1-u))\le e^{-\mu(1-u)}$. Using $\ln x\le x-1$ and $\mathbb E[(M+2)^{-1}]=\int_0^1u\,\mathbb E[u^M]\,du$,
\begin{equation}
 \begin{aligned}
 J(\mu)&\le(\mu+2)\,\mathbb E\!\left[(M+2)^{-1}\right]-1\\
 &\le(\mu+2)\int_0^1u\,e^{-\mu(1-u)}\,du-1\\
 &=\frac{\mu-2+(\mu+2)e^{-\mu}}{\mu^2}
 \le\frac1\mu,
 \end{aligned}
 \label{eq:app_jensen_gap_bound}
\end{equation}
where the last step uses $(\mu+2)e^{-\mu}\le2$. Thus $0\le J(\mu)\le1$ for every $\mu$.

To control the random denominator, let $c_\nu=g(\nu)>0$, and set $R_{\log}=g(M)/g(T)$ on $T>0$ and $R_{\log}=0$ otherwise. Because $M=0$ when $T=0$,
\begin{equation}
 R_{\log}-\frac{g(M)}{c_\nu}
 =R_{\log}\,\frac{c_\nu-g(T)}{c_\nu}.
 \label{eq:app_log_ratio_decomposition}
\end{equation}
On $T\ge\nu/2$, the absolute value of Eq.~\eqref{eq:app_log_ratio_decomposition} is at most $|T-\nu|/[(2+\nu/2)c_\nu]$, using $0\le R_{\log}\le1$ and $g'(x)=1/(2+x)$; on $0<T<\nu/2$, it lies in $[0,1]$. Bounding $\mathbb E|T-\nu|\le\sqrt{\operatorname{Var}(T)}\le\sqrt\nu$ and applying the Bernoulli Chernoff bound $\Pr(T<\nu/2)\le e^{-\nu/8}$ gives
\begin{equation}
 \begin{aligned}
 &\left|\mathbb E[R_{\log}]-\frac{\mathbb E[g(M)]}{c_\nu}\right|\\
 &\qquad\le\frac{\sqrt\nu}{(2+\nu/2)\,c_\nu}+e^{-\nu/8}.
 \end{aligned}
 \label{eq:app_log_denominator_bound}
\end{equation}
Finally, $\mathbb E[g(M)]=g(\mu)-J(\mu)$ and $\mathbb E[R_{\log}\mid T>0]=\mathbb E[R_{\log}]/\omega$ split the error into the denominator term of Eq.~\eqref{eq:app_log_denominator_bound} divided by $\omega$, the Jensen term $J(\mu)/(\omega c_\nu)$, and the normalization term $(1-\omega)g(\mu)/(\omega c_\nu)$. Using $0\le g(\mu)/g(\nu)\le1$, we obtain
\begin{equation}
 \begin{aligned}
 &\left|\mathbb E\!\left[\frac{g(M)}{g(M_{\mathrm{end}})}\middle|M_{\mathrm{end}}>0\right]
       -\frac{g(\mu)}{g(\nu)}\right|\\
 &\quad\le\min\Bigg\{1,
       \frac{J(\mu)+\sqrt\nu/(2+\nu/2)}{\omega g(\nu)}\\
 &\qquad\qquad+\frac{e^{-\nu/8}+1-\omega}{\omega}\Bigg\}.
 \label{eq:qlog_ratio_error_bound}
 \end{aligned}
\end{equation}
This controls the full ratio, including numerator--denominator dependence and valid-sample selection. Since $J(\mu)\le1$ and $1-\omega\le e^{-\nu}$, it is $\mathcal O(1/\ln\nu)$ uniformly in $0\le\mu\le\nu$, and it does not require a large current count. Averaging it under Eq.~\eqref{eq:valid_source_projection_weight} bounds the corresponding curve error.

\paragraph{Scope of the approximations.}
These bounds control the count-mean step within the independent-residual model; they do not bound the additional error from residual correlations, which is assessed empirically in Fig.~\ref{fig:prediction_measurement}(c,d). The ordering $P\le Q_{\log}\le Q$ is exact for the statistics themselves. For the plug-in predictions, the upper comparison persists because concavity of $g$ gives $g(\mu)/g(\nu)\ge\mu/\nu$, but the lower comparison with exact $P$ need not hold.

\subsection{Near-Orthogonal Surrogate}
\label{app:near_orthogonal}

This section derives the near-orthogonal predictions used in Sections~\ref{sec:identity_window} and~\ref{sec:val} and evaluated in Fig.~\ref{fig:prediction_measurement}. Setting $a_{ij}=0$ for all pairs gives $\rho_{ij}=1$ and $C=I$, so $Y,Z_1,\ldots,Z_N$ are independent standard Gaussians, with $N=V-1\ge1$ and $h=\sqrt d/\kappa$. The surrogate retains only the near-orthogonal bulk of pretrained embeddings (Appendix~\ref{app:multimodel_geometry}). We use the superscript $\perp$ for statistics within this model.

\paragraph{Top-1 retention.}
Eq.~\eqref{eq:app_spectrum_approx} becomes
\begin{equation}
 P_{\mathrm{raw}}^{\perp}(\kappa)=\int_{-\infty}^{\infty}\phi(y)\,\Phi(y+h)^N\,dy.
 \label{eq:app_bulk_integral}
\end{equation}
Substituting $u=\Phi(y)$ gives $\int\phi(y)\Phi(y)^N\,dy=1/(N+1)$, so the terminal value is $P_{\mathrm{raw}}^{\perp}(\infty)=1/V$ and, by Eq.~\eqref{eq:app_p_normalization},
\begin{equation}
 P^{\perp}(\kappa)=\frac{(N+1)P_{\mathrm{raw}}^{\perp}(\kappa)-1}{N},
\end{equation}
with $|P^{\perp}-P_{\mathrm{raw}}^{\perp}|\le1/V$.

\paragraph{Rank retention.}
Given $Y=y$, all candidates share the current and terminal overtaking probabilities
\begin{equation}
 p(y,\kappa)=\bar\Phi(y+h),\qquad p_T(y)=\bar\Phi(y),
 \label{eq:near_orthogonal_rank_probabilities}
\end{equation}
so the conditional means are $\mu=Np(y,\kappa)$ and $\nu=Np_T(y)$.
The validity probability and its normalization are
\begin{equation}
 \begin{aligned}
 \omega(y)&=1-\Phi(y)^N,\\
 \mathcal Z_N&=\int\phi(y)\omega(y)\,dy=\frac{N}{N+1}.
 \end{aligned}
 \label{eq:near_orthogonal_valid_weight}
\end{equation}
Since the candidate probabilities no longer depend on the source, the source average in Eq.~\eqref{eq:valid_source_projection_weight} cancels. For finite $y$, $p_T(y)>0$, and conditional on $Y=y$ the nested counts are binomial:
\begin{equation}
 \begin{aligned}
 M_{\mathrm{end}}&\sim\operatorname{Bin}(N,p_T(y)),\\
 M(\gamma)&\sim\operatorname{Bin}\big(\ell,\,p(y,\kappa)/p_T(y)\big),
 \end{aligned}
 \label{eq:near_orthogonal_nested_counts}
\end{equation}
where the second law also conditions on $M_{\mathrm{end}}=\ell\ge1$. Hence $\mathbb E[M(\gamma)/M_{\mathrm{end}}\mid M_{\mathrm{end}}=\ell,Y=y]=p(y,\kappa)/p_T(y)$, and linear retention is exact within this model:
\begin{equation}
 \begin{aligned}
 Q^{\perp}(\kappa)
 &=\frac1{\mathcal Z_N}\int_{-\infty}^{\infty}\phi(y)\omega(y)\\
 &\quad{}\times\left[1-\frac{p(y,\kappa)}{p_T(y)}\right]dy.
 \end{aligned}
 \label{eq:q_near_orthogonal}
\end{equation}
For logarithmic retention, substituting the same conditional means gives
\begin{equation}
 \begin{aligned}
 Q_{\log}^{\perp}(\kappa)
 &\approx\frac1{\mathcal Z_N}\int_{-\infty}^{\infty}\phi(y)\omega(y)\\
 &\quad{}\times\left[1-
 \frac{g(Np(y,\kappa))}{g(Np_T(y))}\right]dy.
 \end{aligned}
 \label{eq:qlog_near_orthogonal}
\end{equation}
This expression retains the count-mean approximation, whose error is bounded by Eq.~\eqref{eq:qlog_ratio_error_bound} with $\mu=Np(y,\kappa)$, $\nu=Np_T(y)$, and weight $\phi(y)\omega(y)/\mathcal Z_N$. All three surrogate statistics depend only on $d$ and $V$ at a given $\kappa$, and decrease from 1 at $\kappa=0$ to 0 as $\kappa\to\infty$.

\subsection{First-Overtaking Scale}
\label{app:first_overtaking}

This section derives Eq.~\eqref{eq:kappa_half} and the corresponding transition width within the near-orthogonal surrogate.

\paragraph{First-overtaking threshold.}
Let $T_N=\max_{1\le j\le N}Z_j$. Top-1 retention ($r=1$) is equivalent to $T_N-Y\le h=1/(\kappa\sigma_b)$, so
\begin{equation}
 \kappa^*=\begin{cases}
 \dfrac1{\sigma_b(T_N-Y)},&T_N>Y,\\[5pt]
 +\infty,&T_N\le Y.
 \end{cases}
 \label{eq:app_first_threshold}
\end{equation}
For $D=T_N-Y$, the density at zero is $f_D(0)=\int_{-\infty}^{\infty} N\phi(t)^2\Phi(t)^{N-1}\,dt>0$. A positive continuous density near zero implies
\begin{equation}
 \begin{aligned}
 &\mathbb E[\kappa^*\mid D>0]\\
 &\quad=\sigma_b^{-1}\mathbb E[D^{-1}\mid D>0]=\infty.
 \end{aligned}
\end{equation}
Even conditioning on a finite threshold does not yield a finite mean, so we characterize the transition by quantiles.

\paragraph{Probit approximation.}
Let $m_N=\mathbb E[T_N]$ and $v_N=\operatorname{Var}(T_N)$. Since $P_{\mathrm{raw}}^{\perp}(\kappa)=\Pr(Y\ge T_N-h)=\mathbb E[\Phi(h-T_N)]$ and $|\frac{d^2}{dt^2}\Phi(h-t)|=|x\phi(x)|_{x=h-t}\le(2\pi e)^{-1/2}$, a second-order Taylor expansion around $m_N$ gives
\begin{equation}
 \left|P_{\mathrm{raw}}^{\perp}(\kappa)-\Phi(h-m_N)\right| \le\frac{v_N}{2\sqrt{2\pi e}}.
 \label{eq:app_probit_error}
\end{equation}
Combined with $|P^{\perp}-P_{\mathrm{raw}}^{\perp}|\le1/V$, this gives Eq.~\eqref{eq:kappa_half}. To leading order $m_N\sim\sqrt{2\ln N}$, but this overestimates $m_N$ at finite $N$, so numerical values below use $m_N$ computed by integration. For $N=V-1=130{,}067$ (Appendix~\ref{app:embedding_validation}), $m_N\approx4.44$ versus $\sqrt{2\ln N}\approx4.85$, and $v_N\approx0.072$, so the bound in Eq.~\eqref{eq:app_probit_error} is below $0.009$.

\paragraph{Location and width.}
Within the surrogate, $P_{\mathrm{raw}}^{\perp}$ decreases strictly and continuously from 1 to $1/(N+1)$, so the quantile $\kappa_p$ defined by $P_{\mathrm{raw}}(\kappa_p)=p$ is unique and finite for $1/(N+1)<p<1$; $N>9$ makes both endpoints of the 90\%--10\% width finite. For $P$ the quantile differs only through the $\mathcal O(1/V)$ normalization. Setting $\Phi(1/(\kappa_p\sigma_b)-m_N)=p$ and $z_p=\Phi^{-1}(p)$ gives
\begin{equation}
 \kappa_p \approx \frac{1}{\sigma_b(m_N + z_p)}=\frac{\sqrt d}{m_N+z_p}.
\end{equation}
At $p=0.5$, $\kappa_{0.5}\approx\sqrt d/m_N\sim\sqrt{d/(2\ln N)}$. With $z=z_{0.9}\approx1.28$, so that $z_{0.1}=-z$, subtracting the two quantiles gives the 90\%--10\% width
\begin{equation}
 W_\kappa=\kappa_{0.1}-\kappa_{0.9}\approx\frac{2z\sqrt d}{m_N^2-z^2}.
\end{equation}
Both location and width scale as $\sqrt d$, while $W_\kappa/\kappa_{0.5}\approx2zm_N/(m_N^2-z^2)\sim2z/m_N$ decreases only as $\mathcal O(1/\sqrt{\ln N})$. Since $\operatorname{Var}(T_N-Y)=v_N+1$ and $v_N\ll1$ for large $N$, the width is set mainly by the source projection $Y$ rather than by fluctuations of the candidate maximum.

\paragraph{Accuracy of the quantiles.}
Write $\varepsilon_N=v_N/(2\sqrt{2\pi e})$. For a finite true quantile $\kappa_p$, Eq.~\eqref{eq:app_probit_error} implies $|\Phi(1/(\kappa_p\sigma_b)-m_N)-p|\le\varepsilon_N$. Write $z_\pm=\Phi^{-1}(p\pm\varepsilon_N)$. If $p\pm\varepsilon_N\in(0,1)$ and $m_N+z_->0$, monotonicity gives
\begin{equation}
 \frac{\sqrt d}{m_N+z_+}\le\kappa_p\le\frac{\sqrt d}{m_N+z_-}.
 \label{eq:app_quantile_interval}
\end{equation}
Width bounds follow by subtracting the endpoint intervals. Since $\varepsilon_N$ depends only on $N$, increasing $d$ alone does not reduce this probability error, and near a zero denominator a small probability error can still produce a large location error. For $d=4096$ and $N=130{,}067$, the probit approximation gives $\kappa_{0.5}\approx14.4$ and $W_\kappa\approx9.1$; Eq.~\eqref{eq:app_quantile_interval} gives $\kappa_{0.5}\in[14.3,14.5]$; and numerical integration of Eq.~\eqref{eq:app_bulk_integral} gives $\kappa_{0.9}\approx11.1$, $\kappa_{0.5}\approx14.4$, $\kappa_{0.1}\approx20.5$, and $W_\kappa\approx9.4$. Initial identity decay thus requires an effective noise magnitude an order of magnitude above 1.

\subsection{Trajectory Factorization and Time Windows}
\label{app:path_factorization}

This section proves Corollary~\ref{cor:path_factorization} and maps the first-overtaking quantiles to trajectory time.

\paragraph{Proof of Corollary~\ref{cor:path_factorization}.}
Under a sample-independent schedule $\kappa(\gamma)$, Proposition~\ref{prop:competition_thresholds} gives each sample's rank as $r_i(\kappa(\gamma))$, which depends on the trajectory only through $\kappa$. Since $\kappa(\gamma)\to\infty$ as $\gamma\to1$, the terminal count is the number of candidates with $b_j>b_i$ (Appendix~\ref{app:threshold_details}), which does not depend on the trajectory; the valid set $E$ is therefore also trajectory-independent. Each per-trajectory score in $P$, $Q$, and $Q_{\log}$ is thus a function of $\kappa(\gamma)$ and the sample, and averaging over sources and noise gives $R(\gamma)=\mathcal R_{\mathcal D}(\kappa(\gamma))$. Consequently, reparameterizing time does not change which competitions occur or their order.

\paragraph{Sample-dependent effective noise magnitudes.}
For spherical interpolation between unit endpoints separated by angle $\theta\in(0,\pi)$, $\kappa(\gamma)=\sin(\gamma\theta)/\sin((1-\gamma)\theta)$. A finite threshold $\kappa^*$, computed with the unit noise $z_S$ of Eq.~\eqref{eq:app_spherical_coupling}, maps exactly to the angle in $(0,\theta)$ given by
\begin{equation}
 \gamma^*=\frac1\theta\operatorname{atan2}\!\left(\kappa^*\sin\theta,\,1+\kappa^*\cos\theta\right).
 \label{eq:app_exact_angle}
\end{equation}
When $\theta=\arccos(u_i^\top z)$ varies across samples, the identity $r(\gamma)=r_i(\kappa_{i,z}(\gamma))$ still holds per sample, but the statistic must be computed by substituting each sample's own $\kappa_{i,z}(\gamma)$ before averaging. The factorization is then approximate. For uniform spherical endpoints, $u_i^\top z$ has mean zero and variance $1/d$, so the angles concentrate near $\pi/2$ and the approximation becomes accurate. For the mask-biased endpoints used by LLaCO, $u_i^\top z$ instead concentrates around $\lambda u_i^\top m/\sqrt{1+\lambda^2}$, so $\theta$ varies with the source token. We therefore measure the scheduling curve directly under the actual endpoint distribution (Appendix~\ref{app:curve_fitting}).

\paragraph{From effective noise magnitude to trajectory time.}
Under the factorization, a $\kappa$-quantile maps to $\gamma_p=\kappa^{-1}(\kappa_p)$, and, to first order in $W_\kappa$, a narrow transition has width $W_\gamma\approx W_\kappa/\kappa'(\gamma_{0.5})$. Monotonicity alone does not compress the transition; compression arises because $\kappa'(\gamma)$ is large where $\kappa$ is large, so the high-$\kappa$ region occupies little trajectory time. On linear trajectories, the fraction of the trajectory with effective noise magnitude above $\kappa$ is $1-\gamma(\kappa)=1/(1+\kappa)$.

\paragraph{Linear trajectories.}
For $s_\gamma=(1-\gamma)u_i+\gamma z$, $\kappa(\gamma)=\gamma/(1-\gamma)$ and $\gamma=\kappa/(1+\kappa)$. Substituting $\kappa_p$ gives
\begin{equation}
 \begin{aligned}
 \gamma_p&=\frac{1}{1+\kappa_p^{-1}}
 \approx\frac{1}{1+\sigma_b(m_N+z_p)},\\
 W_\gamma&\approx\frac{2\sigma_bz_{0.9}}
 {(1+\sigma_bm_N)^2-\sigma_b^2z_{0.9}^2}.
 \end{aligned}
 \label{eq:app_linear_window}
\end{equation}
When $\sigma_b(m_N+z_p)=(m_N+z_p)/\sqrt d\ll1$, a first-order expansion yields
\begin{equation}
 \begin{aligned}
 \gamma_p &\approx 1 - \frac{m_N + z_p}{\sqrt{d}},\\
 W_\gamma&=\gamma_{0.1}-\gamma_{0.9}\approx\frac{2z_{0.9}}{\sqrt d}.
 \end{aligned}
\end{equation}
Top-1 decay therefore occurs in the final $\mathcal O(m_N/\sqrt d)=\mathcal O(\sqrt{\ln N/d})$ fraction of the trajectory, within a window of width $\mathcal O(d^{-1/2})$. For $d=4096$ and $N=130{,}067$, the numerically integrated quantiles give $\gamma_{0.5}\approx0.935$ and $W_\gamma\approx0.036$.

\paragraph{Spherical trajectories.}
For nearly orthogonal endpoints, $\theta\approx\pi/2$ and $\kappa(\gamma)\approx\tan(\pi\gamma/2)$; combined with Eq.~\eqref{eq:app_spherical_sandwich},
\begin{equation}
 \begin{aligned}
 \gamma_p&=\frac2\pi\arctan\kappa_p\approx1-\frac{2}{\pi\kappa_p}\\
 &\approx1-\frac{2\sigma_b(m_N+z_p)}{\pi}.
 \end{aligned}
\end{equation}
Hence $\gamma_{0.5}\approx1-2\sigma_bm_N/\pi$ and $W_\gamma\approx4\sigma_bz_{0.9}/\pi$; with the same $d$ and $N$, $\gamma_{0.5}\approx0.956$ and $W_\gamma\approx0.026$. Spherical interpolation compresses the transition by an additional factor of about $2/\pi$ relative to the linear trajectory.

\subsection{Preservation Bounds and Shared Components}
\label{app:preservation_shared}

This section collects two results that complement Section~\ref{sec:theoretical_analysis}: a sufficient condition for high-probability preservation, and the scaling of thresholds under a shared component. The union bound applies to any endpoint distribution; the closed-form bound requires isotropic Gaussian noise. Exact shared-component scaling requires an equal-coefficient decomposition and consistent source distributions, noise distributions, and amplitude conventions; it requires neither Gaussian noise nor near orthogonality.

\paragraph{High-probability preservation.}
For a given source token, identity decay is the union of candidate overtaking events. Thus, for any endpoint distribution, with $p_{ij}(\kappa)=\Pr[\tau_{ij}<\kappa\mid i]$,
\begin{equation}
\begin{aligned}
 \max_{j\ne i}p_{ij}(\kappa)
 &\le1-P_{\mathrm{raw}}(\kappa\mid i)\\
 &\le\sum_{j\ne i}p_{ij}(\kappa).
\end{aligned}
 \label{eq:identity_union_bound}
\end{equation}
This relation does not require independent overtaking events. Under isotropic Gaussian noise, if every competitor of the given source satisfies $a_{ij}\le a_{\max}<1$, then
\begin{equation}
\begin{aligned}
 &1-P_{\mathrm{raw}}(\kappa\mid i)\\
 &\quad\le(V-1)\bar\Phi\!\left(
 \frac1\kappa\sqrt{\frac{d(1-a_{\max})}{2}}\right).
\end{aligned}
\end{equation}
For $0<\delta<1$ and $\delta/(V-1)<1/2$, define
\begin{equation}
 \kappa_\delta^{\mathrm{bd}}=
 \frac{\sqrt{d(1-a_{\max})/2}}
 {\bar\Phi^{-1}(\delta/(V-1))}.
 \label{eq:preservation_bound}
\end{equation}
Then $\kappa\le\kappa_\delta^{\mathrm{bd}}$ guarantees $P_{\mathrm{raw}}(\kappa\mid i)\ge1-\delta$. If the same similarity bound holds for all sources, the guarantee also applies to source-averaged preservation.

Writing $\phi$ for the standard Gaussian density, integration by parts gives, for $x>0$,
\begin{equation}
 \bar\Phi(x)=\frac{\phi(x)}x-
 \int_x^\infty\frac{\phi(t)}{t^2}\,dt.
\end{equation}
The integral is nonnegative and at most $\bar\Phi(x)/x^2$, yielding
\begin{equation}
 \frac{x}{1+x^2}\phi(x)
 \le\bar\Phi(x)\le\frac{\phi(x)}x.
\end{equation}
Hence $-\ln\bar\Phi(x)=x^2/2+O(\ln x)$, and $\bar\Phi^{-1}(q)\sim\sqrt{2\ln(1/q)}$ as $q\to0$. Substitution gives the leading scale
\begin{equation}
 L_\delta=
 \sqrt{\frac{d(1-a_{\max})}{4\ln((V-1)/\delta)}}\approx\kappa_\delta^{\mathrm{bd}}.
 \label{eq:app_preservation_scale}
\end{equation}
By $\bar\Phi(x)\le e^{-x^2/2}$ for $x\ge0$, this scale is also a conservative sufficient threshold: $L_\delta\le\kappa_\delta^{\mathrm{bd}}$, so $\kappa\le L_\delta$ guarantees preservation at least $1-\delta$. Thus $d(1-a_{\max})\gg4\ln((V-1)/\delta)$ suffices for a large guaranteed preservation scale.
For fixed $a_{\max}$ and $\delta$, this preservation scale grows as $\sqrt d$ with logarithmic dependence on vocabulary size. For example, $d=4096$, $V=130{,}000$, $a_{\max}=0$, and $\delta=0.1$ give $\kappa_\delta^{\mathrm{bd}}\approx9.42$ and $L_\delta\approx8.53$. Retaining each candidate's actual similarity avoids replacing every similarity $a_{ij}$ by a common $a_{\max}$.

\paragraph{Comparison with first-overtaking quantiles.}
A preservation bound gives a sufficient condition for $P_{\mathrm{raw}}(\kappa)\ge1-\delta$, whereas the quantiles of Appendix~\ref{app:first_overtaking} are determined by the preservation curve itself. Let $\kappa_{1-\delta}$ denote the largest $\kappa$ with $P_{\mathrm{raw}}(\kappa)\ge1-\delta$, which exists because $P_{\mathrm{raw}}(\kappa)=\Pr[\kappa^*\ge\kappa]$ is nonincreasing and left-continuous. Under the Gaussian model and a similarity bound holding for every source,
\begin{equation}
 \kappa_\delta^{\mathrm{bd}}\le\kappa_{1-\delta}.
\end{equation}
The left side is a guaranteed preservation range; the right side is the location at which the decay probability reaches $\delta$. For $a_{\max}=0$ and $\delta=0.1$, the near-orthogonal surrogate gives $\kappa_{0.9}\approx11.1$, compared with $\kappa_\delta^{\mathrm{bd}}\approx9.42$. Appendix~\ref{app:empirical_preservation_bounds} compares these bounds with measured first-overtaking quantiles.

\paragraph{Shared components.}
Suppose token directions have an equal-coefficient shared component:
\begin{equation}
 u_j=\sqrt\rho\,c+\sqrt{1-\rho}\,v_j,
\end{equation}
where $0\le\rho<1$, $\|c\|=\|v_j\|=1$, and $c^\top v_j=0$. The residual directions need not be mutually orthogonal. Then
\begin{equation}
\begin{aligned}
 a_{ij}&=\rho+(1-\rho)v_i^\top v_j,\\
 1-a_{ij}&=(1-\rho)(1-v_i^\top v_j),\\
 b_j-b_i&=\sqrt{1-\rho}\,z^\top(v_j-v_i).
\end{aligned}
\end{equation}
For the same noise sample $z$, let $\widetilde\tau_{ij}$ be the competition threshold for the residual directions. Scaling the semantic separation and relative noise advantage gives
\begin{equation}
 \tau_{ij}=\sqrt{1-\rho}\,\widetilde\tau_{ij}.
\end{equation}
The sign of the relative noise advantage is unchanged, so infinite thresholds also correspond. Consequently,
\begin{equation}
\begin{aligned}
 \kappa^*_{i,z}&=\sqrt{1-\rho}\,\widetilde\kappa^*_{i,z},\\
 r_{i,z}(\kappa)&=\widetilde r_{i,z}\!\left(
 \frac\kappa{\sqrt{1-\rho}}\right),\\
 P_{\mathrm{raw}}(\kappa)&=\widetilde P_{\mathrm{raw}}\!\left(
 \frac\kappa{\sqrt{1-\rho}}\right).
\end{aligned}
\end{equation}
Quantiles defined with the same convention therefore satisfy $\kappa_p=\sqrt{1-\rho}\,\widetilde\kappa_p$. In particular,
\begin{equation}
\begin{aligned}
 \kappa_{0.5}&=\sqrt{1-\rho}\,\widetilde\kappa_{0.5},\\
 W_\kappa&=\kappa_{0.1}-\kappa_{0.9}=\sqrt{1-\rho}\,\widetilde W_\kappa.
\end{aligned}
\end{equation}
The location and absolute width scale by the same factor in effective-noise coordinates, leaving $W_\kappa/\kappa_{0.5}$ unchanged. Residual-model preservation bounds scale by the same factor; locations and widths in trajectory time require the corresponding time mapping (Appendix~\ref{app:path_factorization}).

If the residual directions satisfy the near-orthogonal and independent-residual approximations, the predictions of Appendix~\ref{app:first_overtaking} extend to the shared-component model through this scaling. Let $\bar a$ and $\bar a_v$ be mean cosines computed with the same pair weights. Then
\begin{equation}
 \bar a=\rho+(1-\rho)\bar a_v.
\end{equation}
When $\bar a_v\approx0$, we have $\rho\approx\bar a$; in general, their difference is $(1-\rho)\bar a_v$.

\paragraph{Token-dependent coefficients.}
For
\begin{equation}
 u_j=\zeta_jc+\sqrt{1-\zeta_j^2}\,v_j,
\end{equation}
the relative noise advantage contains $(\zeta_j-\zeta_i)z^\top c$, which no longer cancels, so the common scaling is generally no longer exact. Its approximate validity depends on coefficient variation and residual geometry; thresholds are particularly sensitive to projection perturbations when the relative noise advantage is near zero.

\section{Empirical Analysis of Token Geometry and Competition}
\label{app:empirical_statistics}

This section collects measurements of token geometry, validation of competition-statistics predictions, and an empirical comparison of sufficient preservation bounds.

\subsection{Token Geometry Across Pretrained Models}
\label{app:multimodel_geometry}

We measure the original input embedding matrices of the models in Table~\ref{tab:multimodel_geometry}: LLaDA-8B-Instruct~\cite{nie2025large}, LLaDA2.0-mini~\cite{bie2025llada2}, Qwen3-8B~\cite{yang2025qwen3}, Llama-3.1-8B-Instruct~\cite{grattafiori2024llama3}, Dream-v0-Instruct-7B~\cite{ye2025dream}, DeepSeek-R1-0528-Qwen3-8B~\cite{guo2025deepseekr1}, Moonlight-16B-A3B-Instruct~\cite{liu2025muon}, and Nemotron-Labs-Diffusion-8B~\cite{fu2026nemotron}. For each model, we uniformly sample 4,096 source token IDs without replacement from its eligible vocabulary and compare every source with all other eligible candidates. Eligible rows have finite entries and Euclidean norm greater than $10^{-12}$; special tokens declared by the tokenizer or model configuration and rows without a tokenizer entry are excluded. We normalize each original embedding to a unit direction without centering or removing principal components.

Let $a_{ij}=u_i^\top u_j$ for $j\ne i$. We report the mean and standard deviation over evaluated pairs, the fractions with $|a_{ij}|\le0.1$ and $|a_{ij}|\le0.2$, and the mean number of candidates per source with $a_{ij}>0.3$, $0.5$, or $0.7$. For nearest-neighbor statistics, we first compute $m_i=\max_{j\ne i}a_{ij}$ and then report the mean and 50th, 90th, and 99th percentiles of $m_i$ over sources.

\begin{table*}[t]
\centering
\scriptsize
\setlength{\tabcolsep}{2.3pt}
\caption{Directional geometry of pretrained input token embeddings. Each model uses 4,096 uniformly sampled source tokens and its full eligible candidate vocabulary, excluding self-pairs. Coverage uses absolute cosine; candidate counts use signed cosine. Maximum-cosine statistics first take the maximum over candidates for each source, then compute its mean and quantiles across sources. Counts below one are averages over sources.}
\label{tab:multimodel_geometry}
\resizebox{\textwidth}{!}{%
\begin{tabular}{@{}lrrrrrrrrrrr@{}}
\toprule
 & \multicolumn{2}{c}{Pairwise cosine} & Mean max. & \multicolumn{2}{c}{Near-orthogonal pairs (\%)} & \multicolumn{3}{c}{Mean candidate count} & \multicolumn{3}{c}{Maximum cosine quantiles}\\
Model & Mean & Std. & cosine & $|a|\le0.1$ & $|a|\le0.2$ & $a>0.3$ & $a>0.5$ & $a>0.7$ & P50 & P90 & P99\\
\midrule
LLaDA-8B-Instruct & 0.0300 & 0.0415 & 0.4309 & 94.838 & 99.940 & 5.0420 & 0.4033 & 0.0024 & 0.4203 & 0.5854 & 0.6756 \\
LLaDA2.0-mini & 0.0197 & 0.0335 & 0.4687 & 98.299 & 99.960 & 8.0793 & 0.9089 & 0.0481 & 0.4674 & 0.6544 & 0.7469 \\
Qwen3-8B & 0.0172 & 0.0212 & 0.2316 & 99.912 & 99.998 & 0.3860 & 0.0132 & 0.0002 & 0.2117 & 0.3496 & 0.5074 \\
Llama-3.1-8B-Instruct & 0.0161 & 0.0196 & 0.3361 & 99.879 & 99.993 & 1.8340 & 0.2087 & 0.0073 & 0.3177 & 0.5397 & 0.6822 \\
Dream-v0-Instruct-7B & 0.0172 & 0.0336 & 0.3161 & 98.937 & 99.990 & 1.6938 & 0.0496 & 0.0002 & 0.3024 & 0.4480 & 0.5793 \\
DeepSeek-R1-0528-Qwen3-8B & 0.0173 & 0.0213 & 0.2321 & 99.909 & 99.998 & 0.3691 & 0.0098 & 0.0002 & 0.2127 & 0.3561 & 0.4986 \\
Moonlight-16B-A3B-Instruct & 0.0232 & 0.0303 & 0.4940 & 98.804 & 99.964 & 12.2729 & 1.4219 & 0.1504 & 0.4807 & 0.7102 & 0.8069 \\
Nemotron-Labs-Diffusion-8B & 0.0268 & 0.0263 & 0.3453 & 99.142 & 99.989 & 2.1072 & 0.0996 & 0.0000 & 0.3352 & 0.4879 & 0.5968 \\
\bottomrule
\end{tabular}%
}
\end{table*}

Across the measured models, mean pairwise cosine ranges from 0.0161 to 0.0300. Between 94.838\% and 99.912\% of evaluated pairs have absolute cosine at most 0.1, and between 99.940\% and 99.998\% have absolute cosine at most 0.2. At the same time, mean maximum cosine ranges from 0.2316 to 0.4940, and each source has on average approximately 0.37--12.27 candidates with cosine above 0.3. Thus, a near-orthogonal bulk coexists with sparse higher-similarity neighbors.

All geometry computations use FP64. As a spot check, recomputing all candidate cosines of three sources per model without BLAS reproduces every near-orthogonal fraction, candidate count, and nearest neighbor, with absolute differences below $4.0\times10^{-15}$. The survey's exclusion of all declared special tokens leaves 130,061 eligible Nemotron directions. The competition experiment below follows its original ranking vocabulary of 130,068 directions, excluding near-zero rows and \texttt{[MASK]}; these two populations are reported separately.

\subsection{Paired Competition Experiment: Setup and Uncertainty}
\label{app:embedding_validation}

We use the original Nemotron-Labs-Diffusion-8B input embedding tensor with $d=4096$ and $V=130{,}068$ ranking tokens. We sample 4,096 source IDs uniformly without replacement from the ranking vocabulary. We draw 128 fresh independent Gaussian vectors $g\sim\mathcal N(0,I_d/d)$ for every source, yielding 524,288 samples. Each vector is normalized to $n=g/\|g\|$ to obtain its paired uniform spherical endpoint. All four statistics use this common source set and these paired noises; no empirical curve is fitted to obtain the predictions.

Let $M=r-1$ denote the current overtaking count and $M_{\mathrm{end}}$ the terminal overtaking count at the noisy endpoint. Over the common valid population $E=\{M_{\mathrm{end}}>0\}$, we average $\mathbf1\{M=0\}$, $1-M/M_{\mathrm{end}}$, and $1-\ln(1+M/2)/\ln(1+M_{\mathrm{end}}/2)$ to obtain $P$, $Q$, and $Q_{\log}$. Of the sampled trajectories, 524,283 are valid; the remaining five are excluded only from the endpoint-normalized statistics. Raw top-1 retention and expected rank use all 524,288 samples. Definitions and ordering are given in Appendix~\ref{app:rank_normalization}.

Spectrum-aware predictions retain every source--candidate cosine similarity. Gaussian expected rank sums the exact marginal probabilities in Eq.~\eqref{eq:pairwise_prob}, without an independence assumption. Top-1 retention uses the independent-residual spectrum integral in Eq.~\eqref{eq:app_spectrum_approx}. Predictions of $Q$ and $Q_{\log}$ additionally replace conditional counts by their means, as detailed in Appendix~\ref{app:qlog_conditional}. Near-orthogonal predictions set source--candidate similarities to zero under the independent-residual model: linear retention is exact within that surrogate, while logarithmic retention retains the count-mean approximation. Predictions for spherical endpoints use the high-dimensional Gaussian projection approximation in effective-noise coordinates.

We evaluate 1,003 grid positions $q\in[0,1]$ and map them to $\kappa=\tan(\pi q/2)$, including exact clean and noisy endpoints. Measurements retain the sampled noise projections, and spherical trajectory-time measurements additionally use each sample's actual endpoint angle. Normalization, projection, ranking, and prediction calculations use FP64. A total of 3,564 direct score or SLERP checks agree with threshold-derived ranks, and all-trajectory checks verify endpoints, monotonicity, and ordering. Refined numerical integration and direct marginal summation cross-check the predictions.

We estimate pointwise 95\% uncertainty intervals using 1,000 bootstrap replicates clustered by source. Each replicate retains all 128 noises, paired endpoints, and spectrum predictions of every resampled source. Normalized empirical statistics are recomputed from valid-sample sums; theoretical averages use their predicted valid-sample weights. Errors are prediction minus measurement, expressed in percentage points for retention. Expected-rank errors are divided by the measured mean, recomputed in each replicate.

\subsection{Competition Statistics in Effective-Noise Coordinates}
\label{app:competition_comparison}

Fig.~\ref{fig:prediction_measurement} presents the complete set of competition statistics. At a fixed effective noise magnitude, Gaussian and paired spherical endpoints yield nearly overlapping retention and rank curves. Top-1 retention $P$ declines once the first competing candidate overtakes the source token (Fig.~\ref{fig:prediction_measurement}(b)). Logarithmic retention $Q_{\log}$ continues to decrease as the leading ranks deteriorate (Fig.~\ref{fig:prediction_measurement}(c)), whereas linear retention $Q$ responds only to substantially larger cumulative overtaking counts (Fig.~\ref{fig:prediction_measurement}(d)). Expected rank directly records this accumulation (Fig.~\ref{fig:prediction_measurement}(a)). Both the spectrum-aware and near-orthogonal predictions capture these successive stages under both endpoint distributions.

Over the plotted range, the maximum prediction errors for logarithmic and linear retention are approximately 1.30 and 0.60 percentage points, respectively (lower panels of Fig.~\ref{fig:prediction_measurement}(c,d)). During the main decay interval, near-orthogonal predictions lie slightly above measurements, whereas spectrum predictions lie slightly below. The near-orthogonal expected-rank error is concentrated in the early competition region and decreases as the effective noise magnitude increases; retaining the actual spectrum gives closer predictions in this region. Small marginal similarities therefore motivate the bulk approximation without eliminating the effect of the real similarity spectrum.

\begin{table}[!htbp]
\centering
\small
\setlength{\tabcolsep}{5pt}
\caption{Half-retention locations and transition widths in effective-noise coordinates for the common 4,096-source, 128-noise experiment. Pred.\ denotes theoretical predictions; meas.\ denotes measurements with Gaussian or paired spherical endpoints.}
\label{tab:app_joint_kappa_comparison}
\begin{tabular}{@{}lrrr@{}}
\toprule
Estimate & $P$ & $Q_{\log}$ & $Q$\\
\midrule
\multicolumn{4}{@{}l}{\textit{Half-retention effective noise magnitude ($\kappa_{0.5}$)}}\\
\quad Near-orthogonal (pred.) & 14.419 & 23.998 & 91.002\\
\quad Spectrum (pred.) & 13.941 & 23.261 & 88.605\\
\quad Gaussian (meas.) & 14.169 & 23.670 & 89.688\\
\quad Sphere (meas.) & 14.165 & 23.665 & 89.671\\
\midrule
\multicolumn{4}{@{}l}{\textit{90\%--10\% transition width}}\\
\quad Gaussian (meas.) & 9.316 & 64.679 & 514.676\\
\quad Sphere (meas.) & 9.307 & 64.659 & 514.558\\
\bottomrule
\end{tabular}
\end{table}

Table~\ref{tab:app_joint_kappa_comparison} summarizes the separation between competition stages. For Gaussian endpoints, measured half-retention occurs at $\kappa=14.169$ for $P$, $23.670$ for $Q_{\log}$, and $89.688$ for $Q$. Thus, logarithmic retention responds beyond first identity decay while reaching its midpoint well before linear retention.

\subsection{Trajectory Parameterization and Temporal Compression}
\label{app:competition_time_comparison}

Using the paired dataset of Appendix~\ref{app:embedding_validation} and starting from the same unit token direction $u_i$, the linear trajectory $x_\gamma=(1-\gamma)u_i+\gamma g$ has $\kappa=\gamma/(1-\gamma)$. The spherical trajectory $s_\gamma=\operatorname{SLERP}(u_i,n;\gamma)$ has $\kappa=\sin(\gamma\theta)/\sin((1-\gamma)\theta)$ with the actual angle $\theta=\arccos(u_i^\top n)$. Both use cosine ranking. This comparison concerns unit-direction trajectories. Panel (a) of Fig.~\ref{fig:identity_time} instead uses the raw, unnormalized Nemotron-Labs-Diffusion-8B input embeddings: every token in the ranking vocabulary is evaluated at three distinct values of $\gamma$ on a discrete grid, each with an independent endpoint $z\sim\mathcal N(0,I_d/d)$, and top-1 rates under Euclidean and cosine decoding are averaged within each grid value.


Although the paired endpoints give similar curves in effective-noise coordinates, their trajectory parameterizations map these statistics to different time profiles (Fig.~\ref{fig:identity_time}(b, c)). Spherical-trajectory degradation occurs later and over a narrower interval in this experiment. For example, the measured 90\%--10\% width of $Q_{\log}$ is 0.0542 on the linear trajectory and 0.0370 on the spherical trajectory.

\subsection{Preservation Bounds and Observed Transitions}
\label{app:empirical_preservation_bounds}

This comparison uses a sample separate from the paired dataset above, with the same Nemotron-Labs-Diffusion-8B embeddings and ranking vocabulary of 130,068 tokens. We sample 4,096 source tokens uniformly without replacement and draw 128 independent $\mathcal N(0,I/d)$ noises per source, without radial normalization, giving 524,288 samples.

A sufficient preservation bound and a first-overtaking quantile answer different questions: the former guarantees a decay probability no greater than $\delta$, while the latter locates the corresponding transition in the actual distribution. To evaluate the conservatism of union summation while retaining all source--candidate similarities, define
\begin{equation}
 B_{\mathrm{full}}(\kappa)=\frac1S\sum_{i=1}^{S}
 \min\!\left\{1,\sum_{j\ne i}p_{ij}(\kappa)\right\},
 \label{eq:empirical_full_spectrum_bound}
\end{equation}
where $S=4096$ and $p_{ij}$ is given by Eq.~\eqref{eq:pairwise_prob}. This bounds raw identity decay for the equal-weight mixture of the sampled sources, without requiring independent overtaking events.

For comparison, the source-wise maximum similarities have mean $\bar a_{\max}=S^{-1}\sum_i\max_{j\ne i}a_{ij}=0.348411$. Substituting this mean into the maximum-similarity expression gives the heuristic
\begin{equation}
 \widehat\kappa_\delta=
 \frac{\sqrt{d(1-\bar a_{\max})/2}}
 {\bar\Phi^{-1}(\delta/(V-1))}.
 \label{eq:empirical_meanmax_reference}
\end{equation}
Because the mean is not a uniform upper bound on similarity, this heuristic carries no probability guarantee.

\begin{table}[!htbp]
\centering
\small
\setlength{\tabcolsep}{4pt}
\caption{Preservation scales in the separate bound-validation sample. Mean-max denotes the heuristic substitution; Full spectrum solves $B_{\mathrm{full}}(\kappa)=\delta$; Measured denotes the first-overtaking $\delta$ quantile.}
\label{tab:empirical_preservation_bounds}
\begin{tabular}{rrrr}
\toprule
$\delta$ & Mean-max & Full spectrum & Measured\\
\midrule
0.010 & 6.961 & 8.479 & 9.146\\
0.050 & 7.390 & 9.007 & 10.225\\
0.100 & 7.601 & 9.266 & 10.898\\
0.200 & 7.829 & 9.546 & 11.839\\
\bottomrule
\end{tabular}
\end{table}

Table~\ref{tab:empirical_preservation_bounds} reports these scales. At $\delta=0.1$, the full-spectrum union bound gives $\kappa=9.26550$, whereas the measured first-overtaking quantile is $10.89777$. Measured decay at the bound is 1.24035\%, below the allowed 10\%. Thus, union summation remains conservative even when all actual candidate similarities are retained. Fig.~\ref{fig:identity_time}(d) visualizes this comparison across tolerances $\delta$.

\section{Denoising Details}
\label{app:denoising_method}

\subsection{Trajectory and Endpoint Properties}
\label{app:mask_endpoint}

\paragraph{Spherical trajectory and monotonicity.}
For unit vectors $a,b$ at angle $\theta=\arccos(a^\top b)\in(0,\pi)$, spherical linear interpolation traverses the geodesic from $a$ to $b$ at constant speed:
\begin{equation}
 \begin{aligned}
 &\operatorname{SLERP}(a,b;\gamma)\\
 &\quad=\frac{\sin((1-\gamma)\theta)}{\sin\theta}\,a+\frac{\sin(\gamma\theta)}{\sin\theta}\,b.
 \end{aligned}
\label{eq:slerp_def}
\end{equation}
For the noising trajectory in Eq.~\eqref{eq:spherical_path}, $a=u_i$ and $b=z$, so the effective noise magnitude is
\begin{equation}
\begin{aligned}
 \kappa(\gamma)&=\frac{\sin(\gamma\theta)}{\sin((1-\gamma)\theta)},\\
 \kappa'(\gamma)&=\frac{\theta\sin\theta}{\sin^2((1-\gamma)\theta)}>0
\end{aligned}
\end{equation}
for $0\le\gamma<1$. Thus, $\kappa(0)=0$, $\kappa$ is strictly increasing, and $\kappa\to\infty$ as $\gamma\to1$, guaranteeing pathwise rank monotonicity. Because $\theta$ depends on the sample, Proposition~\ref{prop:competition_thresholds} applies per sample; for nearly orthogonal endpoints, $\theta\approx\pi/2$ and $\kappa(\gamma)\approx\tan(\pi\gamma/2)$.

\paragraph{Mean relative noise advantage.}
For $z=\operatorname{normalize}(n+\lambda m)$ with $\lambda>0$, where $n$ is uniform on the unit sphere in $d\ge2$ dimensions and $m$ is the unit mask direction, symmetry of $n$ under orthogonal maps fixing $m$ implies $\mathbb E[z]=c_{d,\lambda}m$ with $c_{d,\lambda}=\mathbb E[z^\top m]$. For $b_j=z^\top u_j$, linearity gives the exact relation
\begin{equation}
 \mathbb E[b_j-b_i]=c_{d,\lambda}\big(m^\top u_j-m^\top u_i\big).
\label{eq:mask_mean_exact}
\end{equation}
Writing $\xi=n^\top m$, outside the probability-zero event $n+\lambda m=0$ we have $z^\top m=(\xi+\lambda)/\sqrt{1+2\lambda\xi+\lambda^2}$. As $d$ increases, $\xi$ concentrates at zero; since this expression is bounded and continuous, $c_{d,\lambda}\to\lambda/\sqrt{1+\lambda^2}$, which gives the mean shift in Eq.~\eqref{eq:mask_mean_shift}:
\begin{equation}
  \begin{aligned}
 \mathbb E[b_j-b_i]&\approx\Delta\mu_j\\
 &=\frac{\lambda}{\sqrt{1+\lambda^2}}\big(m^\top u_j-m^\top u_i\big).
 \end{aligned}
\label{eq:app_mask_mean_shift}
\end{equation}
For $\lambda=1$, used in our method, $z^\top m=\sqrt{(1+\xi)/2}$, so $c_{d,1}=\mathbb E\big[\sqrt{(1+\xi)/2}\big]>0$ and $\Delta\mu_j=(m^\top u_j-m^\top u_i)/\sqrt2$. Candidates with $\Delta\mu_j>0$ are more aligned with the mask direction than the source and receive a positive mean relative noise advantage from the endpoint bias.

\paragraph{Pairwise overtaking under the mask-biased endpoint.}
For $\lambda=1$, the endpoint norm is $\|n+m\|=\sqrt{2(1+\xi)}$ with $\xi=n^\top m=O_p(d^{-1/2})$. Although this norm concentrates at $\sqrt2$, its fluctuation is of the same order as the noise projections and therefore contributes to the variance of the relative noise advantage. A first-order expansion of $(1+\xi)^{-1/2}$ yields
\begin{equation}
 \begin{gathered}
 z^\top u_j=\frac{m^\top u_j+n^\top\tilde u_j}{\sqrt2}+O_p(d^{-1}),\\
 \tilde u_j=u_j-\tfrac12(m^\top u_j)\,m.
 \end{gathered}
 \label{eq:app_effective_codeword}
\end{equation}
To first order, normalization thus acts as a deterministic contraction of each codeword along the mask direction. Writing $A_{ij}=m^\top(u_j-u_i)$, the relative noise advantage decomposes as $b_j-b_i\approx\Delta\mu_j+n^\top(\tilde u_j-\tilde u_i)/\sqrt2$, where the stochastic term is zero-mean, asymptotically Gaussian, and has variance
\begin{equation}
 s_{ij}^2=\frac{\|\tilde u_j-\tilde u_i\|^2}{2d}=\frac{1-a_{ij}}{d}-\frac{3A_{ij}^2}{8d}.
 \label{eq:app_mask_variance}
\end{equation}
The second term constitutes a relative correction of $3A_{ij}^2/[8(1-a_{ij})]$, which is independent of $d$. Approximating the endpoint norm by the constant $\sqrt2$ is therefore valid only when this ratio is small. The pairwise overtaking probability follows as
\begin{equation}
 p_{ij}(\kappa)\approx\bar\Phi\!\left(\frac{(1-a_{ij})/\kappa-\Delta\mu_j}{s_{ij}}\right).
 \label{eq:app_mask_pairwise}
\end{equation}
Within the biased endpoint, $p_{ij}$ increases strictly with $\Delta\mu_j$ for fixed $a_{ij}$ whenever $\kappa>3A_{ij}/(4\sqrt2)$. Since $|A_{ij}|\le2$, this bound remains below $1.1$ and lies well beneath the noise levels at which overtaking occurs. Consequently, among candidates with equal semantic separation from the source, those more aligned with the mask direction have stochastically smaller competition thresholds. Relative to the isotropic endpoint $z=n$, whose variance is $2(1-a_{ij})/d$, the mask bias reduces the variance of $b_j-b_i$ by at least a factor of two, so candidates with $\Delta\mu_j\le0$ overtake later than under the unbiased endpoint. These approximations serve only to interpret token competition; the schedule in Sec.~\ref{sec:semantic_time} is constructed from $Q_{\log}$ measured under the exact endpoint $z=\operatorname{normalize}(n+m)$ (Appendix~\ref{app:curve_fitting}).
By Eq.~\eqref{eq:app_mask_pairwise}, $\Pr[\tau_{ij}\le t_p]=p$ at
\begin{equation}
 t_p=\frac{1-a_{ij}}{\Delta\mu_j-s_{ij}z_p},\qquad\Delta\mu_j>s_{ij}z_p,
 \label{eq:app_mask_threshold_quantile}
\end{equation}
where $z_p=\Phi^{-1}(p)$. In particular, the median threshold is $(1-a_{ij})/\Delta\mu_j$ for $\Delta\mu_j>0$, and the threshold concentrates there when $\Delta\mu_j\gg s_{ij}$.

\paragraph{Conditional thresholds under the mask-biased endpoint.}
For $\lambda=1$, approximating $z\approx(n+m)/\sqrt2$ and defining $Y$ and $Z_j$ from $n$ using the decomposition in Appendix~\ref{app:source_projection}, the approximate overtaking condition is
\begin{equation}
 \begin{gathered}
 Z_j>\rho_{ij}\Big(Y+\frac{\sqrt2}{\kappa\sigma_b}\Big)-\delta_{ij},\\
 \delta_{ij}=\frac{\sqrt{2d}\,\Delta\mu_j}{\sqrt{1-a_{ij}^2}}.
 \end{gathered}
\end{equation}
For uniform spherical $n$, Gaussian projection laws are an additional high-dimensional approximation: at finite $d$, $\operatorname{Var}(Z_j\mid Y=y)=(d-y^2)/(d-1)$, and the residual projections are not independent of $Y$. Within the Gaussian and independent-residual approximations, replace the conditional current and terminal probabilities by
\begin{equation}
 \begin{aligned}
 p_j(y,\kappa)&\approx\bar\Phi\!\left(\rho_{ij}
       \left(y+\frac{\sqrt2}{\kappa\sigma_b}\right)-\delta_{ij}\right),\\
 p_j^T(y)&\approx\bar\Phi(\rho_{ij}y-\delta_{ij}).
 \end{aligned}
\end{equation}
Their sums define the corresponding $\mu$ and $\nu$, and $1-\prod_j(1-p_j^T(y))$ defines the approximate validity weight in Eq.~\eqref{eq:valid_source_projection_weight}. These substitutions give mask-biased approximations of the $Q$ and $Q_{\log}$ predictions in Eq.~\eqref{eq:spectrum_predictions}; the ratio bounds control this surrogate model, not the additional spherical, normalization, or residual-correlation errors. The same substitution in Eq.~\eqref{eq:app_spectrum_approx} gives
\begin{equation}
 \begin{aligned}
 &P_{\mathrm{raw}}(\kappa\mid i)\approx\int_{-\infty}^{\infty}\!\phi(y)\\
 &\quad\times\prod_{j\ne i}\Phi\!\Big(\rho_{ij}\Big(y+\frac{\sqrt2}{\kappa\sigma_b}\Big)-\delta_{ij}\Big)dy,
 \end{aligned}
 \label{eq:app_mask_preservation}
\end{equation}
whose value at $\kappa=\infty$ gives the terminal top-1 retention used in Eq.~\eqref{eq:app_p_normalization}. Since $\cos\theta\approx(\sigma_bY+m^\top u_i)/\sqrt2$, substitute the resulting $\kappa(\gamma,Y)$ before averaging. The pathwise normalization and ordering remain exact for the actual mask-biased endpoint.

\subsection{Retention Curve Fitting}
\label{app:curve_fitting}

\paragraph{Measurement protocol.}
We use the original Nemotron-Labs-Diffusion-8B token embeddings with $d=4096$. Excluding zero-norm embeddings and \texttt{[MASK]} leaves 130,068 tokens, used as both the source population and the ranking vocabulary. Each source token is assigned three distinct positions sampled from the 1,001-point grid $\{0,1/1000,\ldots,1\}$, yielding 390,204 samples and approximately 390 samples per grid point.

Each sample uses an independent endpoint. We normalize a standard Gaussian vector to obtain $n$, then set $z=\operatorname{normalize}(n+m)$. We compute $s_\gamma$ by SLERP and obtain the source token's cosine rank $r(\gamma)$ over the ranking vocabulary. The implementation breaks score ties by token index. Ties between distinct directions occur with probability zero under continuous endpoints, so this agrees almost surely with the strict-overtaking convention of Appendix~\ref{app:threshold_details}. For the same endpoint sample, we also compute $r(1)$ and apply the sample-wise endpoint normalization from Section~\ref{sec:rank_statistics}:
\begin{equation}
\begin{gathered}
 \frac{K_{\log}(\gamma)-K_{\log}(1)}
 {K_{\log}(0)-K_{\log}(1)},\\
 K_{\log}=1-\frac{\ln(1+r)}{\ln(1+V)}.
\end{gathered}
\end{equation}
Samples with a denominator below $10^{-6}$, corresponding here to $r(1)=1$, are excluded. Three samples are excluded, leaving 390,201 valid samples. Their mean within each bin estimates $Q_{\log}(\gamma)$. The linear statistic $Q$ uses sample-wise endpoint normalization as well. In contrast, $P$ is corrected using the terminal top-1 retention rate $p_{\mathrm{end}}$, the fraction of all samples with $r(1)=1$:
\begin{equation}
 P(\gamma)=\frac{P_{\mathrm{raw}}(\gamma)-p_{\mathrm{end}}}{1-p_{\mathrm{end}}}.
\end{equation}

\paragraph{Logistic fit.}
We first fit a decreasing isotonic curve to $Q_{\log}$, weighted by the number of assigned samples per bin, with endpoints fixed at $Q_{\log}(0)=1$ and $Q_{\log}(1)=0$. We then fit the endpoint-normalized Logistic function in Eq.~\eqref{eq:qlog_logistic} by weighted least squares with the same weights. The implementation uses $k=63.7732$ and $\gamma_c=0.966621$, for which $S(1)\approx0.106$. The weighted RMSE relative to the raw bin means and the isotonic curve is approximately 0.0036 and 0.0035, respectively. The fitted normalized curve has a half-retention location of approximately 0.963 and a 90\%--10\% transition width of approximately 0.059.

For the $\lambda=0$ ablation, we repeat the same measurement with the unbiased endpoint $z=n$, yielding 390,202 valid samples, and fit the same family with the same weights. The fitted parameters are $k=106.588$ and $\gamma_c=0.974881$, with weighted RMSE approximately 0.0044 relative to both the raw bin means and the isotonic curve. The fitted normalized curve has a half-retention location of approximately 0.974 and a 90\%--10\% transition width of approximately 0.037; the $\lambda=0$ schedule applies Eq.~\eqref{eq:semantic_time} to this fit.

\paragraph{Alternative fitting families.}
Table~\ref{tab:qlog_fit_families} compares endpoint-constrained families on the same data and reference curve. The endpoint-normalized Logistic has the smallest reported error among these families.

\begin{table}[!htbp]
\centering
\small
\setlength{\tabcolsep}{14pt}
\caption{Reported fitting errors for endpoint-constrained families relative to the isotonic $Q_{\log}$ curve, weighted by bin sample counts.}
\label{tab:qlog_fit_families}
\begin{tabular}{@{}lc@{}}
\toprule
Family & Weighted RMSE\\
\midrule
Endpoint-normalized Logistic & 0.004\\
Stretched exponential & 0.005\\
Beta survival & 0.006\\
Generalized power & 0.007\\
\bottomrule
\end{tabular}
\end{table}
The stretched exponential is
\begin{equation}
 \frac{e^{-c\gamma^p}-e^{-c}}{1-e^{-c}};
\end{equation}
the Beta survival function is $1-I_\gamma(a,b)$, where $I$ is the regularized incomplete beta function; and the generalized power function is $(1-\gamma^a)^b$.

\paragraph{Fits for the $P$ and $Q$ ablations.}
Using the same embeddings, trajectory, and endpoint distribution, we increase the number of assigned positions per source token to ten, yielding 1,300,680 samples. We fit the raw $P$ and $Q$ bin means directly with endpoint-normalized probit functions, weighted by assigned sample counts; isotonic curves serve only as references. For $R\in\{P,Q\}$,
\begin{equation}
\begin{aligned}
 S_R(\gamma)&=\Phi\big(k_R(c_R-\gamma)\big),\\
 R_{\mathrm{fit}}(\gamma)&=\frac{S_R(\gamma)-S_R(1)}
 {S_R(0)-S_R(1)}.
\end{aligned}
\end{equation}
The fitted parameters are $(k_P,c_P)\approx(73.931,0.886)$ and $(k_Q,c_Q)\approx(44.228,1.021)$. Their weighted RMSE values against the respective raw bin means are approximately 0.00277 and 0.00084. Both use the time construction in Eq.~\eqref{eq:semantic_time}, replacing $Q_{\mathrm{fit}}$ with $R_{\mathrm{fit}}$. These ablations therefore share the time construction, but not the fitting family or measurement sample count, with the $Q_{\log}$ schedule.

\paragraph{Inverse mapping.}
We evaluate $t_{\mathrm{sem}}(\gamma)$ on a uniform grid of 10,001 points over $[0,1]$ and invert it by monotone piecewise-linear interpolation. Training and sampling share this table. The endpoints map exactly to $t_{\mathrm{sem}}(0)=0$ and $t_{\mathrm{sem}}(1)=1$.

\subsection{Implementation Details}
\label{app:denoising_implementation}

\paragraph{Time encoder.}
We multiply $t_{\mathrm{sem}}\in[0,1]$ by 1,000 and use 128 frequencies $\omega_i=\exp(-\ln(10^4)i/127)$, $i=0,\ldots,127$, to form 256 sine and cosine features. The condition vector is computed as
\begin{equation}
\begin{aligned}
 a&=\operatorname{SiLU}\big(\operatorname{Linear}_{256\to512}(\mathrm{PE})\big),\\
 c&=\operatorname{SiLU}\big(\operatorname{Linear}_{512\to512}(a)\big).
\end{aligned}
\end{equation}
Thus, the MLP notation in the main text includes the final activation. The layer normalization $\operatorname{LN}(c)$ applied before the layerwise projections is non-affine. Each layer uses a zero-initialized projection from $\mathbb R^{512}$ to $\mathbb R^{6d}$, where $d$ is the hidden dimension, to produce $(\delta,\sigma,g)$ for both attention and feed-forward sublayers. Modulation is applied after the pre-sublayer RMSNorm.

\paragraph{Embedding directional adapter.}
The inverse temperature $\eta$ is parameterized in log space and initialized to 16. The biases $h_j$ are initialized to zero.

\paragraph{Training layout and attention mask.}
Training sequences follow the layout \texttt{[context | clean | noisy]}. The clean segment contains the original target tokens, while the noisy segment contains their continuous states after blockwise noising; both segments share position indices. Context attention is causal. Clean positions attend to the context and causally within the clean segment. Noisy positions attend to the context, clean positions in preceding blocks, and all noisy positions within their own block; they cannot attend to clean positions in the current or subsequent blocks and therefore cannot read the corresponding answers. Each noisy block thus performs bidirectional denoising conditioned on preceding clean blocks, matching blockwise generation, and all target blocks are trained in a single forward pass without sequentially unrolling the blocks.

\paragraph{Time conditioning in training.}
Noisy positions receive the semantic time of their block, while context and clean positions receive zero. Each noisy block independently samples $t_{\mathrm{sem}}$ uniformly from $\{0,1/1000,\ldots,1\}$, maps it to $\gamma$ through the shared inverse mapping, and constructs $s_\gamma$ as in Section~\ref{sec:spherical_noising}.

\paragraph{Blocks and sampling.}
Training uses 8-token blocks; default inference uses 4-token blocks. The sampling grid ends at $t_{\min}=0.01$. Full experimental settings are given in Appendix~\ref{app:training_sampling}.

\subsection{Sampling Procedure}
\label{app:sampling_procedure}

\paragraph{Denoising grid and reverse update.}
Each block of $B$ tokens is initialized at the mask-biased endpoint, $s_0=\operatorname{normalize}(n+\lambda m)$ (trajectory position $\gamma_0=1$), and denoised with $N$ model evaluations on the uniform semantic-time grid
\begin{equation}
  t_k=1-\frac{k}{N-1}(1-t_{\min}),
\label{eq:denoising_grid}
\end{equation}
for $k=0,\ldots,N-1$, with $t_{\min}=0.01$. Inverting Eq.~\eqref{eq:semantic_time} gives $\gamma_k$, so steps concentrate where $Q_{\log}$ changes rapidly. At each intermediate step, the model predicts $p_\theta(\cdot\mid s_k,t_k)$ given the context and preceding blocks; we renormalize the probabilities over the set $\mathcal T_k$ of the $K=512$ highest-probability tokens to $\widetilde p_\theta$, estimate the clean direction, and move along the geodesic:
\begin{equation}
\begin{aligned}
 \widehat x_{0,k}&=\operatorname{normalize}\Big(\textstyle\sum_{j\in\mathcal T_k}\widetilde p_\theta(j)\,u_j\Big),\\
 s_{k+1}&=\operatorname{SLERP}(s_k,\widehat x_{0,k};\varsigma_k),
\end{aligned}
\label{eq:spherical_reverse_update}
\end{equation}
where $\varsigma_k=1-\gamma_{k+1}/\gamma_k$ is the fraction of the arc from $s_k$ to $\widehat x_{0,k}$ traversed in this step. At the final evaluation, we take the argmax token directly, and the accepted tokens serve as clean context for subsequent blocks.

\paragraph{Step indexing.}
The subscript $k$ in Eqs.~\eqref{eq:denoising_grid} and~\eqref{eq:spherical_reverse_update} denotes the inference step, not the trajectory position: $s_0$ corresponds to $\gamma_0=1$, and $\gamma_k$ decreases with $k$. Conditioning on the context and preceding generated blocks is omitted from $p_\theta(\cdot\mid s_k,t_k)$ for brevity. The SLERP update is applied only for $k<N-1$.

\paragraph{Exactness of the reverse update.}
Suppose $s_k$ lies on the true trajectory from the clean direction $x_0$ to the endpoint $z$ at position $\gamma_k>0$, and $\widehat x_{0,k}=x_0$. The arc from $s_k$ to $x_0$ then has length $\gamma_k\theta<\pi$. For $0\le\gamma_{k+1}\le\gamma_k$, the update in Eq.~\eqref{eq:spherical_reverse_update} moves toward $x_0$ along the same geodesic, and the remaining arc length is
\begin{equation}
 \left[1-\left(1-\frac{\gamma_{k+1}}{\gamma_k}\right)\right]
 \gamma_k\theta=\gamma_{k+1}\theta.
\end{equation}
Hence, under these assumptions, the update reaches the true trajectory state at $\gamma_{k+1}$ exactly.

\paragraph{Key--value caching.}
After a block is decoded, we compute layerwise keys and values from the clean embeddings of the accepted discrete tokens and cache them before generating the next block, matching the clean-prefix conditioning used in training.

\section{Training and Sampling Details}
\label{app:training_sampling}

\subsection{Training and Sampling Settings}
\label{app:training_sampling_settings}

\paragraph{Initialization and normalization.}
The backbone and output head are initialized from the original pretrained Nemotron-Labs-Diffusion-8B weights; the embedding directional adapter, time encoder, and timestep-conditioned modulation modules are newly initialized. Token embeddings and their unit directions remain fixed, while all other parameters are trained. The time-encoder output uses non-affine LayerNorm, and the backbone retains its native RMSNorm.

\paragraph{Optimization.}
We use Adam with $\beta_1=0.9$, $\beta_2=0.95$, $\epsilon=10^{-8}$, weight decay 0.1, and a gradient clipping threshold of 1.0. Peak learning rates are $10^{-6}$ for the backbone and output head and $10^{-4}$ for newly introduced modules, followed by cosine decay after 500 warmup steps. The global batch size is 8. Training runs for 165k optimization steps and processes 1.32M training examples. This amounts to approximately 1.685B text tokens, including context, of which approximately 1.300B target tokens, including EOS padding, are supervised. These figures exclude the pretraining of the backbone.

\paragraph{Data processing and mixture.}
Training examples are length-filtered, deduplicated, and checked for valid tokens. Dialogue, mathematics, science, and code are mixed in proportions of 30\%/30\%/30\%/10\% by example count. Context and target lengths are capped at 2,048 and 4,096 tokens, respectively. Training uses 8-token blocks. The final block of each example is filled with EOS tokens, which participate in noising and supervision; padding added outside examples to form a batch is excluded from the loss.

\paragraph{Licenses and ethical considerations.}
Both the backbone~\cite{fu2026nemotron} and the training data are publicly available, and we use them in accordance with their respective licenses. Since our contributions concern the corruption and denoising processes of continuous diffusion rather than new data sources or capabilities, LLaCO introduces no additional ethical risks beyond those of its backbone and training data.

\paragraph{Sampling settings.}
Default inference uses 4-token blocks and 4 model evaluations per block. We construct a uniform semantic-time grid from $t_{\mathrm{sem}}=1$ to $t_{\min}=0.01$ and invert the shared mapping to obtain geometric positions. Intermediate steps estimate the clean direction from the 512 highest-probability tokens. The final evaluation directly takes the argmax vocabulary prediction without an additional endpoint decode.

\subsection{Model Landscape and Baseline Selection}
\label{app:baseline_selection}

Table~\ref{tab:model_landscape} categorizes representative large language models by generative paradigm. Autoregressive models factorize generation into sequential next-token predictions under a causal attention mask~\cite{grattafiori2024llama3,yang2025qwen3,olmo2025olmo3,liu2026ministral3}. Discrete diffusion models generate text via iterative denoising over discrete state spaces~\cite{nie2025large,ye2025dream}, whereas continuous diffusion models operate over continuous embedding spaces~\cite{tae2025tess2,chen2026langflow,hu2026elf}. Nemotron-Labs-Diffusion~\cite{fu2026nemotron} unifies autoregressive, diffusion, and self-speculative decoding within a single architecture, and thus constitutes a distinct autoregressive--diffusion hybrid that cannot be simply categorized as either an autoregressive or a diffusion model.

\begin{table}[t]
\centering
\small
\setlength{\tabcolsep}{3pt}
\caption{Representative language models grouped by generative paradigm. Params denotes the reported model scales. ``Compared'' indicates inclusion in Table~\ref{tab:overall_ablation}; instruction-tuned variants are evaluated where available.}
\label{tab:model_landscape}
\begin{adjustbox}{max width=\columnwidth}
\begin{tabular}{@{}llll@{}}
\toprule
Model & Params & Initialization & Compared \\
\midrule
\multicolumn{4}{@{}l}{\textit{Autoregressive}} \\
Llama~3~\cite{grattafiori2024llama3} & 8B--405B & From scratch & No \\
Qwen3~\cite{yang2025qwen3} & 0.6B--235B & From scratch & No \\
Olmo~3~\cite{olmo2025olmo3} & 7B/32B & From scratch & No \\
Ministral~3~\cite{liu2026ministral3} & 3B/8B/14B & Pruning + distillation & No \\
\midrule
\multicolumn{4}{@{}l}{\textit{Autoregressive--diffusion hybrid}} \\
Nemotron-Labs-Diffusion~\cite{fu2026nemotron} & 3B/8B/14B & Joint AR--diffusion & No \\
\midrule
\multicolumn{4}{@{}l}{\textit{Discrete diffusion}} \\
LLaDA~\cite{nie2025large} & 8B & From scratch & Yes \\
LLaDA-1.5~\cite{zhu2025llada15} & 8B & LLaDA-8B-Instruct & Yes \\
Dream~\cite{ye2025dream} & 7B & Qwen2.5-7B & Yes \\
\midrule
\multicolumn{4}{@{}l}{\textit{Continuous diffusion}} \\
TESS-2~\cite{tae2025tess2} & 7B & Mistral-7B & Yes \\
DSL-LLaDA~\cite{yu2026dslllada} & 8B & LLaDA-8B-Instruct & Yes \\
LangFlow~\cite{chen2026langflow} & 0.13B & From scratch & No \\
ELF~\cite{hu2026elf} & 0.1B--0.65B & From scratch & No \\
RePlaid~\cite{yang2026replaid} & 0.07B--1B & From scratch & No \\
Cola DLM~\cite{guo2026continuous} & $\sim$2B & From scratch & No \\
\textbf{LLaCO} & 8B & Nemotron-Labs-Diffusion-8B & -- \\
\bottomrule
\end{tabular}
\end{adjustbox}
\end{table}

\paragraph{Scope of comparison.}
As a continuous diffusion model, LLaCO is compared with discrete and continuous diffusion models of the same scale, both of which generate text via iterative denoising. Autoregressive and hybrid models are listed for completeness and excluded from the comparison.

\paragraph{Discrete diffusion baselines.}
We compare against LLaDA-8B-Instruct, LLaDA-1.5-8B, and Dream-v0-Instruct-7B, all of which are instruction-tuned models of the same scale as LLaCO.

\paragraph{Continuous diffusion baselines.}
TESS-2 and DSL-LLaDA adapt 7--8B pretrained language models to continuous denoising with publicly released checkpoints, and both are evaluated under the same protocol as LLaCO. Other recent continuous diffusion models are excluded for the following reasons.
LangFlow, ELF, and RePlaid are trained from scratch at 0.07B--1B parameters on pretraining corpora such as LM1B, OpenWebText, and SlimPajama without instruction tuning, and are primarily evaluated by perplexity, generative perplexity, or conditional generation. Evaluating them on instruction-following benchmarks would conflate disparities in model scale and training data with differences in generative paradigm.
Cola DLM is a continuous latent diffusion model with approximately 2B parameters. Its released checkpoint is research-oriented and not instruction-tuned, and its official pipeline targets an eight-task zero-shot suite (LAMBADA, MMLU, OBQA, HellaSwag, RACE, SIQA, SQuAD, and Story Cloze) under a dedicated generative protocol. The release further notes that the open-source implementation differs slightly from the internal implementation used in the paper. Since our benchmarks fall outside this released evaluation setting, a faithful assessment under our protocol is not supported, and Cola DLM is therefore excluded.

\subsection{Block Size and Sampling Steps}
\label{app:block_steps}

We compare generation quality under different inference block sizes $B$ and numbers of model evaluations per block $N$. We first vary the block size with $N=B\in\{4,8,16\}$, and then fix $B=8$ and vary $N\in\{4,8,16\}$. All settings use the default LLaCO-8B checkpoint with other decoding settings held fixed, and $(4,4)$ is the default configuration. Results are reported in Table~\ref{tab:block_steps}.


\paragraph{Block size.}
In the left group, $N$ scales with $B$, so this comparison mainly reflects the effect of block size. Quality decreases markedly as $B$ grows: from $(4,4)$ to $(16,16)$, GSM8K drops from 80.74 to 48.07 and MATH from 62.12 to 24.28, while HumanEval and MBPP lose about 33 and 26 points, respectively. Notably, $(4,4)$ also outperforms $(8,8)$, whose block size matches the 8-token blocks used in training, so the advantage of smaller inference blocks cannot be attributed to a train--inference mismatch in block size.

\paragraph{Steps per block.}
At $B=8$, increasing $N$ from 4 to 8 improves all four tasks, while increasing it further to 16 leads to slight declines on all but HumanEval. Overall, GSM8K and MATH vary by less than 3.7 points across $N\in\{4,8,16\}$, far less than the variation induced by block size.

\paragraph{Discussion.}
The effect of the number of steps can be understood from two aspects. First, the $Q_{\log}$ schedule concentrates evaluations where token identity is decided. The fitted retention curve changes mainly within a narrow window around $\gamma\approx0.96$ (Appendix~\ref{app:curve_fitting}); with $N=4$, the two intermediate evaluations already fall in this window ($\gamma\approx0.971$ and $0.949$), and larger $N$ mostly subdivides the same interval, which limits the overall variation. Second, additional steps allow the model to re-estimate the clean direction at more intermediate states along the denoising trajectory. When the current state lies on the noising trajectory and the estimated clean direction coincides with the true one, the geodesic update in Eq.~\eqref{eq:spherical_reverse_update} reaches the true trajectory state at $\gamma_{k+1}$ exactly (Appendix~\ref{app:sampling_procedure}). In practice, however, the estimate $\widehat x_{0,k}$ depends on both $s_k$ and $t_k$, so this per-step geometric exactness does not preclude step-size errors in the overall sampling procedure. As shown in Table~\ref{tab:block_steps}, increasing $N$ from 4 to 8 improves performance on all four tasks, whereas a further increase to 16 yields task-dependent changes. We hypothesize that this pattern reflects a trade-off between more frequent refinement of the clean-direction estimate and trajectory deviations introduced by inaccurate estimates during the updates.

Within a block, the estimates for all positions are formed simultaneously and committed together at the final evaluation. Dependencies among these positions must therefore be resolved through the evolution of the continuous states rather than by conditioning on accepted tokens. Larger blocks place more mutually dependent tokens in this joint decision, which is consistent with the strong effect of $B$ and with the block-size trend reported for discrete block diffusion~\cite{arriola2025block}.

\end{document}